\documentclass[12pt,journal,onecolumn]{IEEEtran}
\ifCLASSINFOpdf
\else
\fi
\usepackage{caption}
\usepackage[export]{adjustbox}

\usepackage{graphicx}
\graphicspath{{fig/}}

\usepackage{multirow}

\usepackage{epstopdf}
\usepackage{hyperref}

\usepackage{dblfloatfix}
\usepackage{afterpage}

\usepackage{textcomp}

\usepackage{comment}

\usepackage{cite}
 
\usepackage{subfig,color}
\usepackage{tablefootnote}

\usepackage[cmex10]{amsmath}
\usepackage{amssymb}

\usepackage{float}

\usepackage{xspace} 

\usepackage{soul}
\usepackage{array}
\newcolumntype{L}[1]{>{\raggedright\let\newline\\\arraybackslash\hspace{0pt}}m{#1}}
\newcolumntype{C}[1]{>{\centering\let\newline\\\arraybackslash\hspace{0pt}}m{#1}}
\newcolumntype{R}[1]{>{\raggedleft\let\newline\\\arraybackslash\hspace{0pt}}m{#1}}

\usepackage[percent]{overpic}

\usepackage[linesnumbered,ruled,norelsize]{algorithm2e}

\makeatletter
\newcommand{\removelatexerror}{\let\@latex@error\@gobble}
\makeatother

\usepackage{multirow}

\usepackage{etoolbox}
\makeatletter
\patchcmd{\@makecaption}
  {\scshape}
  {}
  {}
  {}
\makeatletter
\patchcmd{\@makecaption}
  {\\}
  {:\ }
  {}
  {}
\makeatother

\usepackage{color}

\usepackage{booktabs,caption,fixltx2e}
\usepackage[flushleft]{threeparttable}

\begin{document}
%
\title{Super-Resolution via Deep Learning}
%
%
%

\author{
Khizar~Hayat~\IEEEmembership{Member,~IEEE}.
\thanks{The author Khizar Hayat was associated with both COMSATS Institute of IT (CIIT) Abbottabad, Pakistan and University of Nizwa, Oman. (Email:khizar.hayat@unizwa.edu.om).}}

\IEEEtitleabstractindextext{%
\begin{abstract}
The recent phenomenal interest in convolutional neural networks (CNNs) must have made it inevitable for the super-resolution (SR) community to explore its potential. The response has been immense and in the last three years, since the advent of the pioneering work, there appeared too many works not to warrant a comprehensive survey. This paper surveys the SR literature in the context of deep learning. We focus on the three important aspects of multimedia - namely image, video and multi-dimensions, especially depth maps. In each case, first relevant benchmarks are introduced in the form of datasets and state of the art SR methods, excluding deep learning. Next is a detailed analysis of the individual works, each including a short description of the method and a critique of the results with special reference to the benchmarking done. This is followed by minimum overall benchmarking in the form of comparison on some common dataset, while relying on the results reported in various works.   
\end{abstract}

}

\maketitle

\IEEEdisplaynontitleabstractindextext

%
\IEEEpeerreviewmaketitle

\section{Introduction}\label{sec:introduction}
\IEEEPARstart {F}{or} the last decade or so, interest in deep learning has skyrocketed. Deep learning had been around for many years but got little interest from researchers; may be due to low computing powers and network speeds or may be huge unstructured data were not commonplace? Now everybody is talking about it and want to contribute in the form of tutorials, special courses, articles, blogs, source codes, APIs and original research. And that has helped a lot in developing a general understanding about the underlying concepts. One can't recall any other issue got that much help available in such a short span of time, be it iPython notebooks on GitHub, Matlab implementations (MatConvNet~\cite{ Vedaldi2015}), C++ API in the form of Caffe~\cite{Jia2014} or R-language based source codes or even patents.~\cite{Yin2016}. If you are interested, there's no dearth of resources and they will make sure you learn it.  

As far as super-resolution is concerned, the pioneering work on the role of deep learning is as fresh as 2014~\cite{Dong2014}. Since then, there has been a mushroom growth and several works have appeared focusing not only images but also videos and higher dimensional multimedia data, especially depth maps or range images, digital elevation models (DEMs) and multispectral images. The cornerstone of all the relevant research is the single image super-resolution method~\cite{Dong2016}, called Super-Resolution Convolutional Neural Network (SRCNN), which is an extension of the pioneering work~\cite{Dong2014} called SRCNN-Ex by its authors. The importance of SRCNN can be gauged by the fact that since its appearance, you will hardly find a super-resolution work not using it as one of its benchmarks. 

In this paper we attempt to survey the deep learning literature in the context multimedia super-resolution. The main focus is on three areas, \emph{viz.} still images, videos and higher dimensions, especially the range data. For each of the three, we first introduce the relevant benchmarks before reviewing the contemporary literature on deep learning based super-resolution, which is followed by a comparison of important methods on the basis of a common dataset. In case of higher dimensions, however, a common dataset is elusive. The description of benchmarks is in the form of publicly available datasets and important super-resolution methods that are not deep learning based; the latter will be introduced in the subsequent discussion, none the less. For benchmarking among methods, we rely on the results reported in the literature and have therefore compiled the tabular data from relevant resources. 

The rest of the paper is arranged as follows. Section~\ref{sec:back} presents essential background concepts with reference to deep learning and super-resolution. Section~\ref{sec:img} surveys the contemporary deep learning works on image SR. A special part of this section concerns the importance of SRCNN with references from literature. The next two sections, i.e. Section~\ref{sec:VSR} and Section~\ref{sec:3D}, follow the same approach for videos and 3D/depth maps, respectively. Section~\ref{sec:concl} concludes the paper.
\section{Background}\label{sec:back}
Before visiting the deep learning based super-resolution literature, it is expedient to give a brief description of the background concepts needed for the understanding of this work.
\subsection{Deep Learning}
Classical machine learning (ML) techniques are characterized by the application of the underlying algorithm to a select group of features extracted after a laborious and intelligent processing and pre-processing. The key is to fine tune and select the best features which is normally time consuming and not possible without considerable expertise and knowledge of the domain. These techniques are thus handicapped by their limited ability to process raw natural data; such data are always huge~\cite{LeCun2015a}. 
Representation learning pertains to the auto-discovery of representations from input raw data. "Deep-learning methods are representation-learning methods with multiple levels of representation having simple but non-linear modules" to get "higher and abstract representation". The goal is to use the composition of enough such transformations in order to learn very complex functions. The peculiarities are, a) the non-involvement of humans in designing layers of features and b) employment of a generic learning procedure to get features from data. "Deep learning allows computational models that are composed of multiple processing layers to learn representations of data with multiple levels of abstraction ... Deep learning discovers intricate structure in large data sets by using the backpropagation algorithm to indicate how a machine should change its internal parameters that are used to compute the representation in each layer from the representation in the previous layer"~\cite{LeCun2015}.

\begin{figure}[htbp]
\centering
\includegraphics[width=15cm]{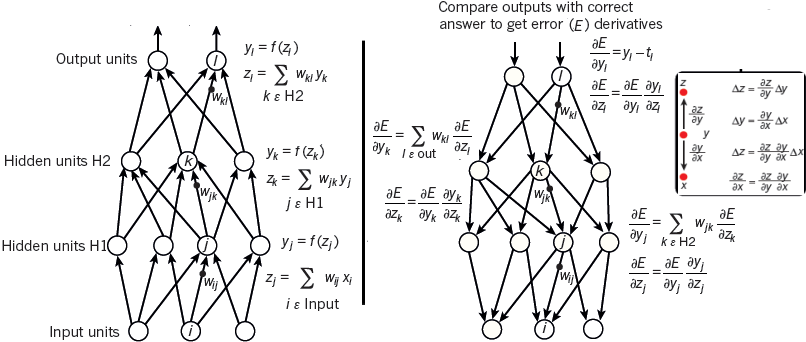}\\
\caption{Backpropagation (after~\cite{LeCun2015}).}
\label{fig:bp}
\end{figure}

According to Hinton~\cite{Hinton2012}
, the use of backpropagation algorithm for learning multiple layers of non-linear features can be traced back to 1970’s. It, however, did not get attention of ML community for its perceived poor exploitation of multiple hidden layers, time-delay and convolutional nets being the exceptions; in addition, its performance was poor in recurrent networks. It was employed with supervised approaches and was thus aptly criticized for its slowness, requirement of labeled data and the high chance of getting stuck in poor local optima. By 2006~\cite{Hinton2006}, with the advent of high processing speeds, these limitations of backpropagation were overcome by using unsupervised learning~\cite{Golovko2016} in place of supervised one and thus applying it directly on raw rather than the labeled data. The idea is to "keep the efficiency and simplicity of using a gradient method for adjusting the weights, but use it for modeling the structure of the sensory input." In other words, "adjust the weights to maximize the probability that a generative model would have produced the sensory input", i.e. learn $p(data)$ not $p(label|data)$. As can be observed from Fig.~\ref{fig:bp}, backpropagation is just a practical application of the chain rule of derivatives. 

The power of neural networks (NNs) lies in their ability to approximate any continuous function~\cite{CS231n}
, as demonstrated in~\cite{Cybenko1989,Nielsen2017}. A $n$-layer NN is characterized by an input layer and $n-1$ hidden layers. In a fully connected version, neurons between two adjacent layers are fully connected, but neurons within a single layer share no connections. The forward pass of a fully-connected layer corresponds to one matrix multiplication followed by a bias offset and an activation function. The commonly used activation functions are, sigmoid, $tanh$ and the Rectified Linear Unit (ReLU~\cite{Glorot2011}). Parameter initializations for the first forward pass of a NN can be carried out in a variety of ways (like all zero weights, random weights etc). The outputs of each forward pass, during the training phase, are compared with the ground truth to get the error signal. Backpropagation of this error signal yields the derivative/gradient for learning that serves to readjust the weights of the parameters for the next forward pass. The process repeats itself till acceptable level of convergence whereby the optimized parameters should ideally classify each subsequent test case correctly. 

The birth of Convolutional neural networks (CNN) or ConvNets can be traced back to 1988~\cite{LeCun1990}\footnote{A claim contested by Schmidhuber~\cite{Schmidhuber2015} who traces it to as early as 1965.} wherein backpropagation was employed to train a NN to classify handwritten digits. Subsequent works by LeCun evolved into what was later known as LeNet5~\cite{LeCun1998}. After that there's virtual lull till late noughties~\cite{Culurciello2017} when GPUs were efficient enough to culminate in the work~\cite{Ciresan2010}. Since then a floodgate has opened and we hear of various architectures in the form of AlexNet~\cite{Krizhevsky2012}, ZFNet~\cite{Zeiler2014}, GoogLeNet~\cite{Szegedy2015} DenseNet~\cite{Huang2016a} etc.; for a detailed overview one can consult~\cite{Culurciello2017,Canziani2016}. 

\begin{figure}[htbp]
\centering
\includegraphics[width=15cm]{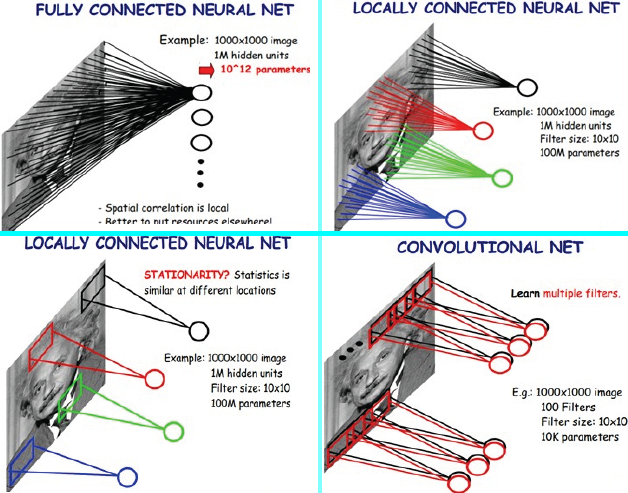}\\
\caption{CNN as seen from NN perspective~\cite{LeCun2015a}.}
\label{fig:cnn}
\end{figure}
\begin{figure}[htbp]
\centering
\includegraphics[width=15cm]{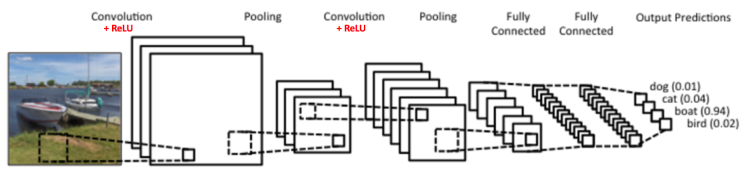}\\
\caption{A Simplified CNN.}
\label{fig:simpleCNN}
\end{figure}

The metamorphosis from fully connected NN to locally connected NN to CNN is illustrated in Fig.~\ref{fig:cnn}. As can be seen, rather than being fully connected, the CNN employs convolutions leading to local connections, where each local region of the input is connected to a neuron in the output. The input to a CNN is in the form of multiple arrays, such as a color image with three 2D arrays (length $\times$ width) in accordance to RGB or YCbCr channels. The number of channels is called depth and constitutes the 3rd D; note that more than three channels are not uncommon, e.g with hyperspectral images. A CNN is made up of Layers with each layer transforming an input 3D volume to an output 3D volume~\cite{CS231n}, typically, via four distinct operations~\cite{ujjwalkarn2016}, \emph{viz.} convolution, a non-linear activation function (ReLU), pooling or sub-sampling and classification (fully connected Layer). A simplified CNN is illustrated in Fig.~\ref{fig:simpleCNN}\footnote{https://www.clarifai.com/technology}. A CNN can be described as several convolution layers with nonlinear activation functions (e.g. ReLU or sigmoid) applied to each layer. Each convolution layer applies several (may be thousands) distinct filters\footnote{The filter is applied in spatial domain using the usual sliding window approach, typical with digital images, and hence the name convolution.} (also called feature maps) and combines their results. These filters are automatically learnt during the training part based on the task in hand, e.g. if the task is image classification the learning concerns, a) detecting edges from raw pixels in the first layer, b) then use the edges to detect simple shapes in the second layer, c) followed by the use of these shapes in higher layers to detect higher-level features (such as facial shapes) and d) using the latter in the last layer for classification~\cite{Britz2015}. 
%
For further details on CNN, readers are encouraged to consult~\cite{LeCun2015,CS231n,ujjwalkarn2016,Britz2015,Culurciello2017,Canziani2016,Shaikh2017,OlahBlog}.

\subsection{Super-Resolution Basics}\label{SR}
In fields - like astronomy, remote sensing, microscopy and tomography etc. - the acquired images may be handicapped by a variety of factors. These may include imperfections of measuring devices - like optical degradations or limited capacity of the sensors - and instability of the observed scene - object motion or media turbulence. The affected images may be indistinct, noisy and deficient in spatial and/or temporal resolution~\cite{Cristobal2008}. The remedy could be one or both of blind deconvolution (to remove blur) and super-resolution. 

\begin{figure}[htbp]
\centering
\includegraphics[width=15cm]{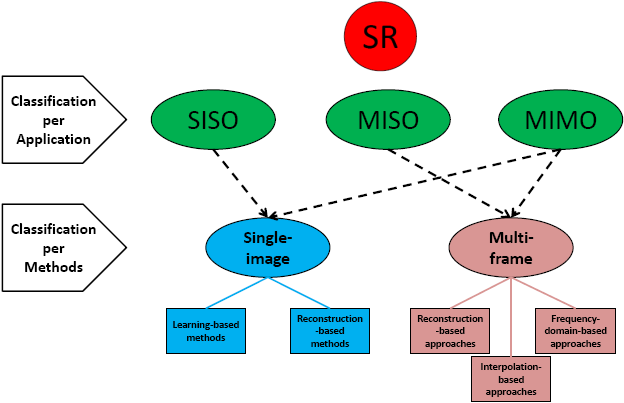}\\
\caption{Taxonomy of SR techniques~\cite{Bevilacqua2014}.}
\label{fig:SRClassify}
\end{figure}

Super-resolution (SR) refers to an estimation of high resolution (HR) image/video from one or more low resolution (LR) observations~\cite{Nasrollahi2014} of the same scene, usually employing digital image processing and ML techniques. Being an inverse problem in most cases, there may be more than one solution, each requiring the construction of a forward observation model~\cite{Bevilacqua2014}. Probably, the first effort on the subject can be traced back to as early as 1984~\cite{Huang1984}; the explicit use of the term 'super-resolution' was a bit later in 1990~\cite{Irani1990}.
In~\cite{Nasrollahi2014}, a very detailed taxonomy is given but for the sake of brevity it is better to rely on the three tier classification of~\cite{Bevilacqua2014} given in Fig~\ref{fig:SRClassify}. The first tier classifies on the basis of both input and output, as single input single output (SISO), multiple input single output (MISO) and multiple input multiple output (MIMO). MIMO pertains to video SR and can be easily merged with the second one, which makes the first tier redundant. Hence it is better to directly classify according to the second tier, i.e.  
into two main categories, namely \emph{single image} super-resolution (SISR) and multiple image or \emph{multi-frame} super-resolution. From here onwards the rest of the Section~\ref{SR} is almost entirely based on~\cite{Bevilacqua2014}.

SISR concerns the problem of estimating an underlying HR image, given a single LR image of the scene, under the assumption that the original imaging set up is not available. Being an ill-posed problem, since several HR may correspond to the input LR image, SISR can be likened to ordinary "analytical" interpolation - like linear, bicubic, and cubic
splines. The task may thus be to compute the missing pixel intensities in the HR grid as averages
of known pixels, which may work well in smooth parts but wrought with dangers in case of discontinuities, in the form of edges and corners, that may lead to ringing and blurring artifacts. Hence more sophisticated insight, in addition to interpolation, is needed to super-resolve the input. There are two types of SISR algorithms:
\begin{enumerate}
	\item \textbf{Learning methods} employ ML techniques to locally estimate the HR details of the output image. These may be \emph{pixel-based}, involving statistical learning~\cite{He2011,Zhang2012}, or \emph{patch-based} involving dictionary based LR to HR correspondence of squared pixel blocks (called patches). The latter ones, also called \emph{example-based} methods~\cite{Freeman2002,Glasner2009}, exploit internal similarities within the same image and may adopt various approaches, e.g. neighbor embedding~\cite{Chan2009}, sparse coding~\cite{Yang2010}, or a blend of these~\cite{Gao2012}. 
	\item \textbf{Reconstruction methods} usually require explicit prior information (in the form of distribution, energy function etc.) while defining constraints for the target HR image.  This may be carried out in a variety of ways, like sharpening of edge details~\cite{Dai2007}, regularization~\cite{Aly2005} or deconvolution~\cite{Shan2008}.
\end{enumerate}
Some methods~\cite{Dong2013} may be termed as a melange of ML and reconstruction methods. Note that most "of the recent SISR methods fall into the example based methods which try to learn prior knowledge from LR and HR pairs, thus alleviating the ill-posedness of SISR. Representative methods include neighbor embedding regression~\cite{Chang2004,Timofte2013,Timofte2015}, random forest~\cite{Schulter2015,Salvador2015} and deep convolutional neural network(CNN)~\cite{Dong2014,Dong2016,Liang2016,Kim2016}"~\cite{Liang2017}.

In multi-frame SR, the input usually consists of more than one LR images, usually from slightly different perspectives of the scene in question. It is assumed that each input image is a degraded version of an underlying HR scene spoiled by blurring, down-sampling and affine transforms. For the latter case, integral shifts are considered trivial, carrying no useful information; ideally, fractional or sub-pixel shifts have greater information value. There are three types of multi-frame methods:
\begin{enumerate}
	\item \textbf{Interpolation methods}~\cite{Bose2006,Patti1997} usually consisting of three steps, namely registration, interpolation and deblurring. 
	\item \textbf{Frequency-domain methods}~\cite{Ji2009,Chappalli2005} gather disparate clues about high frequencies of the underlying HR from the DFT, DCT, DWT or any other frequency domain representation of the input LR frames. Due to their localized nature, DWT domain is better suited.
	\item \textbf{Regularization methods}~\cite{Tian2010,Belekos2010} are useful in case of limited number of LR images or ill-conditioned blur operators, and try to use either deterministic or stochastic regularization strategy to incorporate some prior knowledge of the unknown HR image. 
\end{enumerate}
For a detailed treatment on the subject, the readers are recommended to consult~\cite{Nasrollahi2014,Luong2009,Bevilacqua2014,Tian2011,Huang2015a,Karimi2014,Cristobal2008}.
\section{Deep Networks for Image Super-resolution}\label{sec:img}
According to~\cite{Kim2016a}, while employing a CNN for restoration tasks (like SR and denoising), pooling or subsampling may be counter-productive as important image details may be discarded. Hence pooling layers are usually avoided in SR tasks which again has its downside; each additional convolutional layer means a new weight layer and hence more parameters with the consequences in the form of overfitting and too huge a model to store and retrieve.
\subsection{Image Benchmarks}
\subsubsection{Image databases}
\begin{table*} [htb]
\renewcommand\tabcolsep{2pt}
\caption{Publicly Available Image datasets.}
\label{table:image_datasets}
\centering
\begin{tabular}{|c|l|p{13cm}|}
\hline
 \textbf{S/No.} & \textbf{Name with reference} & \textbf{Details}\\ 
\hline 
 1 & ImageNet~\cite{Russakovsky2015} & The detection dataset by ILSVRC, which consists of around 400,000 images.\\
 2 & Timofte dataset~\cite{Timofte2015} & Widely used by SISR researchers, it consists of 91 training images~\cite{Yang2010} and two test datasets; Set5 and Set14 with 5 and 14 images, respectively.\\
3 & Berkeley segmentation dataset~\cite{Martin2001} & BSD300 and BSD500 - 200 images for testing  (B[SD]200), the rest for training and validation. \\
4 & CIFAR-10~\cite{Krizhevsky2009} & After the Canadian Institute for Advanced Research - 6000 images in each of 10 classes of which 5000 are in training set and the rest in test set. CIFAR-100: 600 images in each of 100 classes.\\
5 & L20~\cite{Timofte2016} & Has very large images, between 3m pixels to up to 29m pixels, while the other datasets have images below 0.5m pixels~\cite{Timofte2014supp}.\\
6 & The General-100 dataset~\cite{Dong2016a} & Contains 100 bmp-format images (with no compression), since the proponents  believe the BSD500 are not optimal for SR tasks, due to JPEG format. The image sizes range from 710x704 (large) to 131x112 (small). They are all of good quality with clear edges but fewer smooth regions (e.g., sky and ocean), thus are very suitable for the SR training.\\
7 & Urban 100~\cite{Huang2015b} & Containing 100 HR images with a variety of real-world structures; famous for its self-similarities.\\
8 & The Kodak PhotoCD dataset~\cite{KodakPCD} & Consists of 24 lossless true color images without compression artifacts, and is used as a standard testing set for many image processing works~\cite{Salvador2015}.\\
9 & The super texture dataset~\cite{Dai2015} & Provides 136 texture images.\\
10 & 291 from ~\cite{Schulter2015} & A combination of S/No. 2 above and BSD200\\ 
11 & MNIST~\cite{LeCun1998a} & Modified National Institute of Standards and Technology database: a large database of handwritten digits.\\
& - Binary version~\cite{Salakhutdinov2008} & \\
& - MNIST corners dataset by~\cite{Dahl2017} & Constructed by randomly placing an MNIST digit in either the top-left or bottom-right corner \\
12 & CelebA~\cite{Liu2015} & Centrally cropped faces \\
13 & LSUN Bedrooms~\cite{Yu2015} & Bedroom images \\
14 & The van Hateren dataset~\cite{vanHateren1998} & 4167 gray scale images of 12 bit depth (mostly nature or buildings) with 1536 X 1024 pixels each and a gray scale. \\
15 & MS-COCO~\cite{Lin2014} &  91 easily recognizable objects types - a total of 2.5 million labeled instances in 328k images. \\
16 & YFCC100M~\cite{Thomee2015} &	The Yahoo Flickr Creative Commons 100 Million Dataset containing 100 million media objects (around 99.2 million photos and 0.8 million videos) under a Creative Commons license.\\
17 & LIVE~\cite{Sheikh20xx} &  A database variously distorted images with accompanied subjective assessments from human observers. The images were originally acquired for a project on generic shape matching and recognition.\\
18 & Manga109~\cite{Fujimoto2016} & A publicly available dataset of 109 Japanese comic books with numerous comic sketches.\\
\hline
\end{tabular}
\end{table*}
In Table~\ref{table:image_datasets}, we list many of the image datasets that are popular with the SR community. Some of the datasets have already been partitioned, by their proponents, to training/validation and testing sets; there is, however, no hard or soft rule and many works use them without restricting to these partitions. Sometimes the researchers improvise on the datasets, e.g. in~\cite{Dong2016}, the 91 Timofte dataset is decomposed into 24,800 sub-images for training along with 395,909 images (over 5 million sub-images) from ImageNet. Similarly some authors combine more than one datasets, e.g. training 291 (91 Timofte +200BSD) from~\cite{Schulter2015} is a popular choice for training~\cite{Kim2016,Yang2017}. 
\subsubsection{Non-CNN super-resolution methods for benchmarking}
An attempt~\cite{Yang2014} at benchmarking the state of the art SISR methods, focusing mainly on~\cite{Glasner2009,Irani1991,Kim2010,Yang2010,Yang2013,Timofte2013,Sun2008} in addition to bicubic interpolation, stops short of giving any verdict because of the mercurial performance of the methods on BSD200 and LIVE1 dataset~\cite{Sheikh2005}. For obvious reasons, no CNN based method is there; indirect comparison may still be possible as most of the included references have also been employed for the benchmarking of the CNN based methods appeared since. In this section we attempt to describe the non-CNN methods mainly used for benchmarking emerging methods. The major CNN based benchmarking methods are presented anyhow in the next sections.

Following are some of the principal non-CNN methods favored for contemporary benchmarking in the image super-resolution literature:
\begin{enumerate}
	\item \textbf{Bicubic interpolation:} one of the widely used classical interpolation methods; others being nearest neighbor and bilinear.
	\item \textbf{NE+~\cite{Chang2004}:} a set of neighbor embedding methods that selects several LR candidate patches in the dictionary by using a nearest neighbor search and employs their HR version for the reconstruction of HR output patches;  may be via least squares (NE+LS) or local linear embedding (NE+LLE) or even Non-Negative Least Squares (NE+NNLS~\cite{Bevilacqua2012}).
	\item \textbf{SC or SrSC~\cite{Yang2010}:} finds the sparse representation to sparsely approximate the input LR patch and then applies the resultant coefficients to sparsely generate the corresponding HR output patch.
	\item \textbf{KK~\cite{Kim2010}:} directly learns a map from input LR images to target HR images based on kernel ridge regression (KRR) using a sparse approach that combines kernel matching pursuit and gradient descent.
	\item \textbf{K-SVD~\cite{Zeyde2012}:} refers to a combination of K-SVD (from~\cite{Aharon2006}) and Orthogonal Matching Pursuit (OMP) for efficient dictionary learning in order to improve upon the sparse method SC.
	\item \textbf{A+~\cite{Timofte2015}, ANR and GR~\cite{Timofte2013}:} Anchored Neighborhood Regression (ANR) is an effort to improve upon K-SVD and SC by introducing a ridge regression (solvable offline and storable per dictionary atom/anchor).  A less accurate but more efficient variation employs global regression; hence the name GR. A+ (advanced ANR) is a later improvement that, unlike ANR, does not solely learn from dictionary atoms, but involves all the training patches in the local anchor neighborhood. Although having similar time complexities, A+ has been shown by the authors outperforming ANR and GR.
	\item \textbf{Self-Ex~\cite{Huang2015b}:} self-similarity algorithm that includes rubber-band transformations, after estimating the expected deformation of recurring patches, in order to expand the internal patch search space. The method outperforms state of the art methods, especially A+, on a synthetic database developed by the authors.
	\item \textbf{SRF~\cite{Schulter2015}:} (super-resolution forests) relies on direct mapping from LR to HR patches using random forests (RFs).  The authors demonstrate a relation of contemporary SISR to locally linear regression and try to fit in RFs into this framework. The method has many variants in the form of RF linear (RFL), RFL+ and its advanced version ARFL+ (also called ASRF).
	\item \textbf{NBRSF~\cite{Salvador2015}:} employs a \emph{Na\"ive-Bayes SR forest} with bimodal trees for example-based SR using a hierarchical external learning strategy, that provides a fast local linearization search, followed by a fast Local Naive Bayes strategy for patch-wise tree selection.
	\item \textbf{IA~\cite{Timofte2016}:} or improved A+ is the result of a generic 7-way strategy proposed by its authors for the amelioration of any given SR method. 
\end{enumerate}
The last three methods are important as they are part of a counter-narrative against CNN strategies and one can find SRCNN in their comparisons.
\subsection{State of the Art Methods on Image SR}
\begin{figure}[htbp]
\centering
\includegraphics[width=15cm]{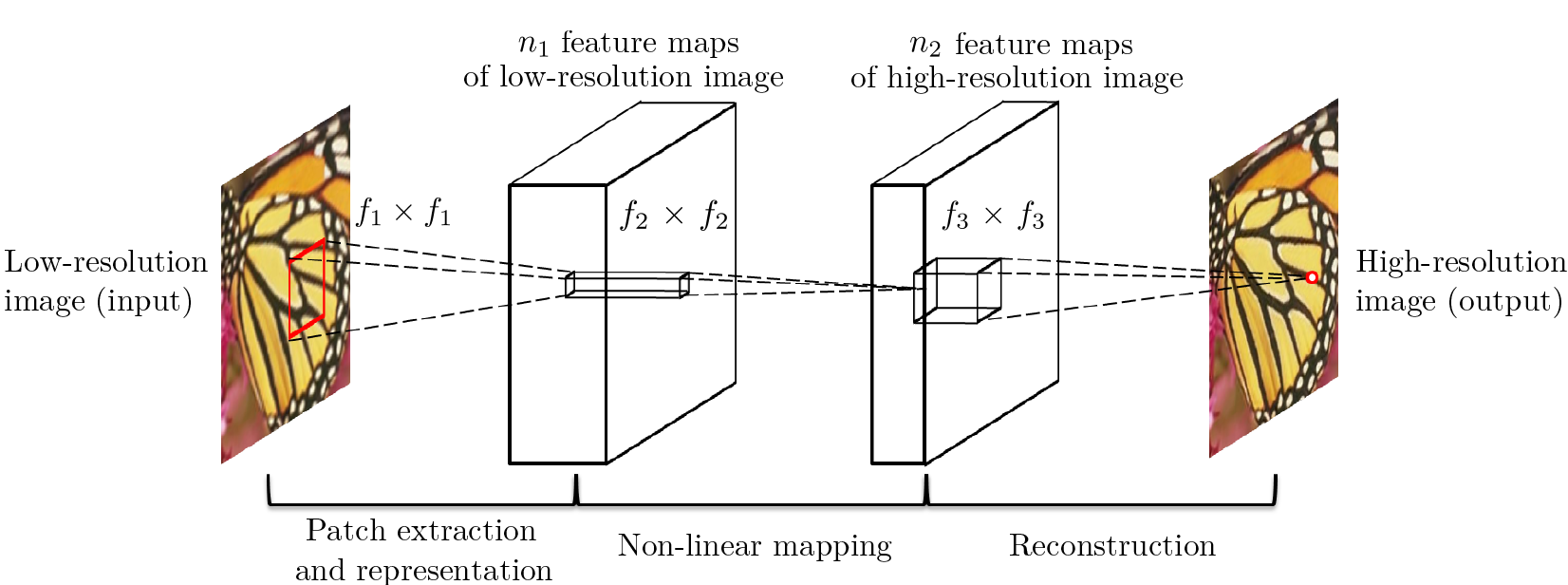}\\
\caption{SRCNN~\cite{Dong2016}.}
\label{fig:SRCNN}
\end{figure}

The SISR method~\cite{Dong2016}, called SRCNN\footnote{http://mmlab.ie.cuhk.edu.hk/projects/SRCNN.html}, is illustrated in Fig.~\ref{fig:SRCNN}. The said work is an extension of an earlier work~\cite{Dong2014}\footnote{In literature, SRCNN refers to~\cite{Dong2016}- the earlier version~\cite{Dong2014} is referred to as SRCNN-Ex by its authors.} after the introduction of larger filter sizes and additional mapping layers. The authors rely on learning directly an end-to-end mapping, between the input LR image and the corresponding HR output image, that is represented as a deep CNN. The method uses a bicubic interpolation as its pre-processing step followed by the extraction of overlapping patches, via convolution, as high dimensional vectors with as many feature maps as their dimensions. The vectors are then non-linearly mapped to each other and subsequently aggregated in the form of patches to get the reconstructed HR image that is supposed to be as close to the ground truth as possible. The authors boast results comparable to ANR~\cite{Timofte2013} and "somehow to" KK~\cite{Kim2010}. However, it is reported in~\cite{Huang2015b} that on Urban100 dataset (and even with BSD100), SRCNN is outperformed by methods like Self-Ex~\cite{Huang2015b}, A+~\cite{Timofte2015} and KK, by a fair margin. 
The authors of~\cite{Dong2016} view CNN as an extension to sparse-coding-based SR methods~\cite{Yang2010} but for the fact that the former jointly optimizes all layers unlike the segregation of components in the latter. 

SRCNN "has only convolutional layers which has the advantage that the input images can be of any size and the algorithm is not patch-based."~\cite{Kappeler2016}. Although SRCNN claims efficiency in view of what the authors call a lightweight structure, it's still a far cry. With that being said an effort to improve the efficiency is in the offing in the form of fast SRCNN (FSRCNN)~\cite{Dong2016a} but a look at the preliminary version has a lot to answer especially about the quantification of speed improvements. In addition, the proposed FSRCNN\footnote{http://mmlab.ie.cuhk.edu.hk/projects/FSRCNN.html} improves upon the original in terms of PSNR. 
For the sake of efficiency, the FSRCNN~\cite{Dong2016a} proposal replaces pre-processing bicubic interpolation step of SRCNN by a post-processing (fifth) step in the form of deconvolution. Other than that, the pipeline has four convolution layers in the form of feature extraction, shrinking, mapping and expanding. That is, the mapping is preceded by the shrinking feature dimensions and followed by expanding back. Moreover, the filter sizes are proposed to be reduced again but with the introduction of additional mapping layers. The author claims 40 times improvement which is debatable in the absence of theoretical analysis.

%
\begin{figure}[htbp]
\centering
\includegraphics[width=15cm]{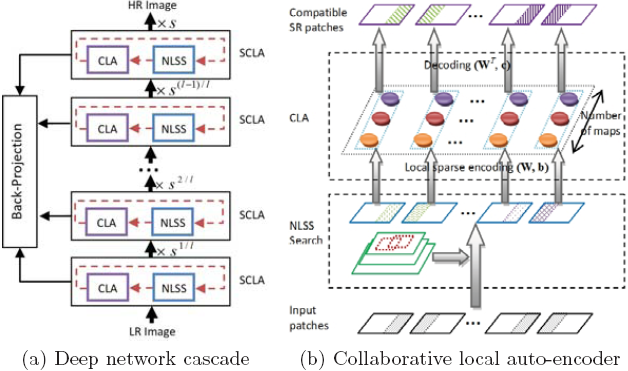}\\
\caption{DNC~\cite{Cui2014}.}
\label{fig:DNC}
\end{figure}
A deep learning method, called deep network cascade (DNC~\cite{Cui2014}), gradually upscales the input LR, layer by layer with a small scale factor, to eventually get the HR image. The successive refinements are based on searching non-local self-similarities in order to enhance the high-frequency details of the patches to which the image is partitioned in a way similar to the internal example-based method~\cite{Glasner2009}, as pointed out in~\cite{Greaves2016}. The patches are then fed to each layer of a cascading multi-stacked network of collaborative local auto-encoder (CLA) for noise suppression and collaborating the conformance among overlapping patches enhanced by the aforementioned process. The scheme in outlined in Fig.~\ref{fig:DNC}. The stacked CLA (SCLA) is a concatenation to incorporate multiple models into the cascade for better SR.
The reported results demonstrate the superiority of DNC in comparison of various state of the art methods  on sparse coding. The reported set of images for experimentation is small, however; very few images\footnote{The authors do refer http://vipl.ict.ac.cn/paperpage/DNC/ for detailed results but the link had been inaccessible, at least at the time of writing this article.}.

\begin{figure}[htbp]
\centering
\includegraphics[width=15cm]{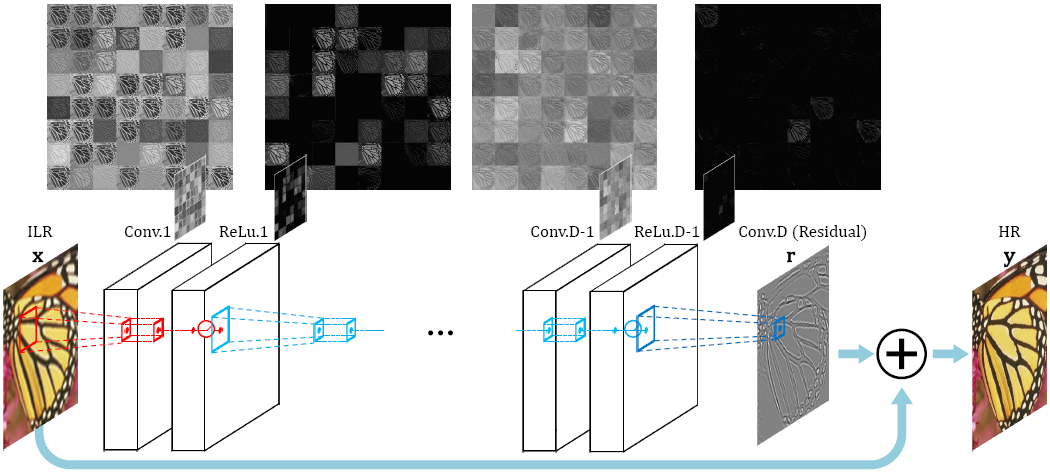}\\
\caption{VDSR~\cite{Kim2016}.}
\label{fig:VDSR}
\end{figure}
For obvious reason the deeper the CNN, more accurate should be results. The VDSR\footnote{may stand for Very Deep Super-Resolution, as the authors do not mention it}~\cite{Kim2016} method\footnote{http://cv.snu.ac.kr/research/VDSR/}, illustrated in Fig.~\ref{fig:VDSR} uses a very deep convolution network inspired by VGG-net used for ImageNet classification~\cite{simonyan2014}. Using a depth of 20 weight layers in a cascaded deep network, involving small filters multiple times, the authors report efficiency in exploiting context information over large image regions. The slow convergence issue, related to high depth, is tackled by using very high learning rates; the downside is gradient exploding which is mitigated by learning residuals only and adaptive gradient clipping. The work is also extended to multiscale SR using a single network. 
The authors are so enthusiastic about their results that they choose the very first figure in their article to demonstrate the superiority of VDSR par rapport the state of the art, both in terms of efficiency and visual quality; especially SRCNN which is shown to be outperformed by 0.87dB~\cite{Knoche2017}. Moreover, further reported results show that VDSR outperforms five state of the art reference methods, with SRCNN finishing second in terms of PSNR and A+ finishing second in terms of time efficiency.

Deeper networks may lead to high accuracies but can bring two issues to the fore, \emph{viz.} overfitting and huge model. With that in view, Kim \emph{et al.}~\cite{Kim2016a} propose 
what they call deeply-recursive convolutional network (DRCN), to apply the same convolutional layer recursively as many as 16 times. The idea is to avoid introducing additional parameters while having an increased depth. The authors deal with training problems, especially the exploding/vanishing gradients, by adopting the recursive-supervision strategy of~\cite{Lee2015} and skipping of layers from~\cite{Shelhamer2017}. 
The authors use the same simulation environment as they have used for VDSR~\cite{Kim2016}. Although they have not included VDSR as a reference, a comparison is still possible as the datasets and the reference methods are the same. The reported results are comparable to VDSR, and even outperforms the latter in case of Set5, as far as final image quality is concerned. The authors, however, conveniently ignore the time efficiency which was important in the context of the article in hand. 

The authors in~\cite{Wang2015} argue that despite the emerging popularity of data driven approaches, the sparse coding paradigm has not lost its value for its domain knowledge and can be potentially very beneficial if combined with deep learning, especially if embedded in a cascaded structure. This, they say, may not only lead to efficiency and better training but also a reduced model size. Their method, the sparse coding based network (SCN), exploits NN approximation of sparse coding in the form of the learned iterative shrinkage and thresholding algorithm (LISTA) proposed in~\cite{Gregor2010}.  SCN - later extended to~\cite{Liu2016} - is claimed to have been compact, accurate and imperceptible in relation to SRCNN. The authors also propose a cascaded version (CSCN) of their method, that employs multiple SCNs, with better artifact reduction and scaling flexibility.
The reported results are better than those by SRCNN  and A+, in terms of PSNR, SSIM, subjective perception and time efficiency. With the latter, however, the results are empirical and no theoretical analysis is carried out.

\begin{figure}[htbp]
\centering
\includegraphics[width=15cm]{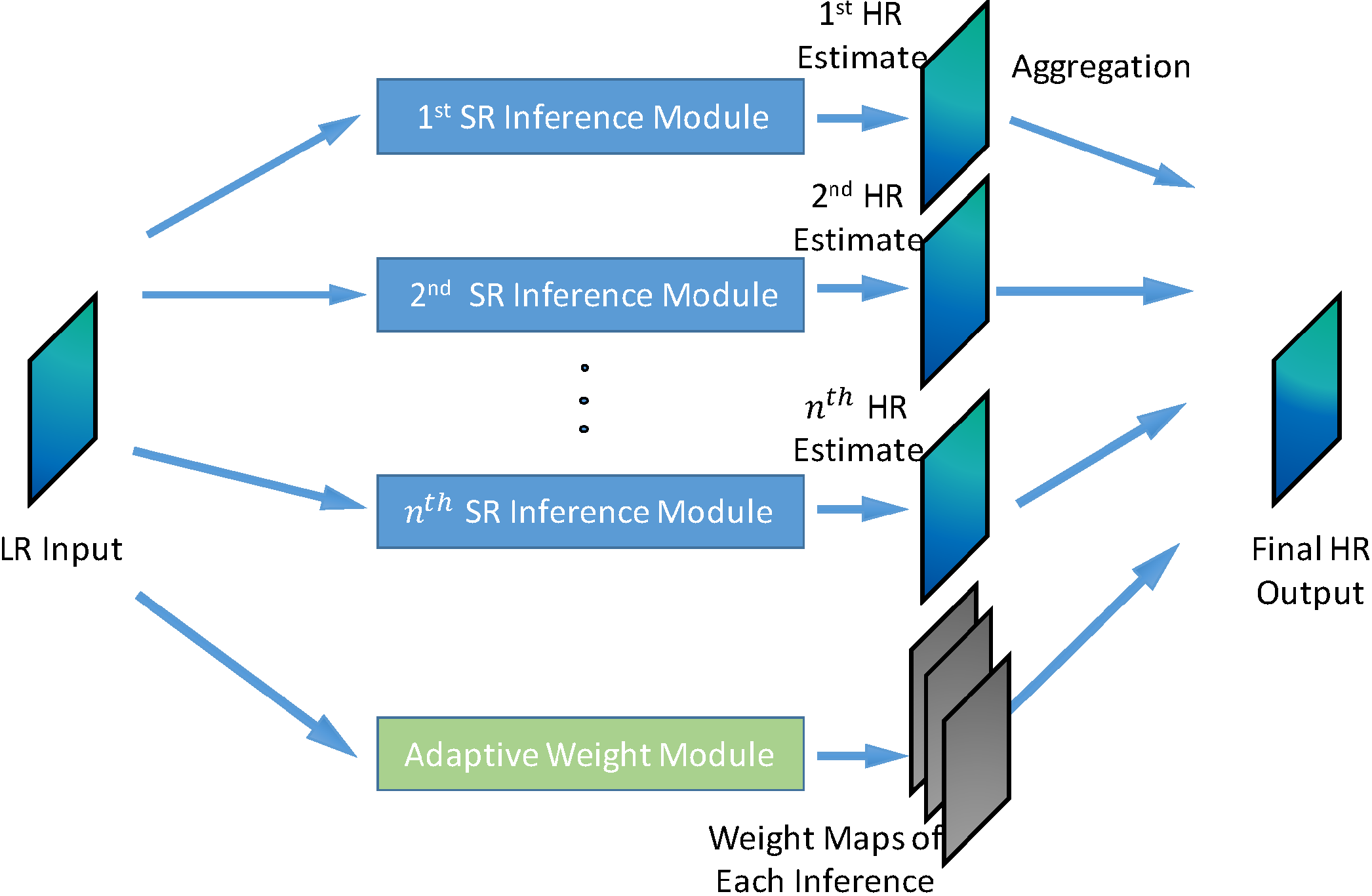}\\
\caption{MSCN~\cite{Liu2017}.}
\label{fig:MSCN}
\end{figure}

In another yet to be properly published work~\cite{Liu2017}, the authors apply a number of SR methods to the LR image, independently, to get various HR estimates which are then combined, on the basis of adaptive weights, to get the final result~\ref{fig:MSCN}. The authors refer to their method as MSCN-$n$, where n is the number of employed  inference modules. 
The reported results demonstrate superiority of MSCN to the state of the art methods - like A+, SCN and SRCNN - as far as image quality is concerned. The method is also shown to be efficient with a comparable performance to~\cite{Wang2015}. 

Another wok that exploits various image priors during the training phase of a deep CNN~\cite{Liang2016} is called SCRNN-Pr. One aspect of prior information focuses edge/texture restoration and the other concentrates on gradual upscaling  via parallel structure recurrence. The authors claim efficient training speed along with better image quality par rapport the state of the art. Strangely enough, despite the claim of SRCNN-Pr, the same first author has not even alluded to SRCNN-Pr as a reference method, in his later related works ~\cite{Liang2017,Liang2017a}; all the other state of the art method, he relied on for comparison, are there.

The work in~\cite{Yang2017}, which was later extended in~\cite{Liang2017} to include some more related reference methods, employs the concepts of residual networks~\cite{He2016}, for SISR. Skip connections are used to counter exploding/vanishing of gradients and a parameter economic CNN (width, depth and skip connections) is proposed for faster training. In essence they propose two schemes, namely R-basic and R-deep, with the former having 22 convolutional layers with 0.3M parameters while the latter having 34 convolutional layers with 5M parameters. They have chosen half a dozen reference method but their main focus is on comparison with a 20-layer VDSR, involving 0.7M parameters, which is showing better results than other reference and R-basic methods but is outperformed by R-Deep. Both R-basic and R-deep are demonstrated to converge faster than VDSR, in terms of the number of epochs; without highlighting the per epoch time complexity, however.

With robustness and efficiency in perspective, the Deep Projection CNN (DPN) method in~\cite{Liang2017a} employs what the authors call model adaptation to exploit the repetitive structures in the LR image. The DPN has three parts: feature extraction/representation via stacked CNN/ReLU layers, inference via CNN/ReLU based projections, and CNN only HR reconstruction. The authors claim up to 0.3dB PSNR gain with a 40-layer DPN (and 0.7M parameters) over reference methods, like A+, SRCNN and VDSR (same 20 layers and 0.7M parameters, as in~\cite {Liang2017}). The reported results are almost the same as reported~\cite{Liang2017} and the latter, despite being from the same first author, does not appear on the reference list. In addition, no time efficiency results are reported.

The pixel recursive super-resolution network in~\cite{Dahl2017} is composed of a conditioning network and a prior network. Whilst the former is a CNN that converts the input LR image to logits to predict the log-likelihood of each HR pixel, the latter is a PixelCNN~\cite{Oord2016}. The probability distribution is represented as a softmax operator applied to the sum of the output logits of the two networks. The model thus "synthesizes realistic details into images while enhancing their resolution".  The PixelCNN  is by itself a CNN variant of Pixel Recurrent Neural Network (PixelRNN)~\cite{Oord2016a}. The authors claim success in the form of subjective quality perception of the resultant images but the method is outperformed by one of their reference methods~\cite{He2016} in terms of quantitative measures, like PSNR. Strangely enough, the method is not compared with any CNN based method.

In.~\cite{Wang2015a}, the authors propose a model named deep joint super resolution (DJSR) in order "to adapt deep model for joint similarities". The authors argue that the CNN "model has a clear analogy to classical sparse coding methods", like~\cite{Yang2010}. While using the Stacked Denoising Convolutional Auto-Encoder (SDCAE\footnote{An implementation: https://github.com/ifp-uiuc/anna}), reported in~\cite{Masci2011}
, the method takes randomly corrupted LR images as input to output HR images by combining the auto-encoders and CNNs (even SRCNN may be possible).  SDCAE pre-trained on external examples followed by refinement with reliable multi-scale self examples. 
The reported results demonstrate comparable performance to SRCNN and better than A+, DNC etc.

The authors in~\cite{Mao2016} term their method very deep Residual Encoder-Decoder Network - “RED-Net” for short  - that consists of a series of convolutional and subsequent deconvolutional layers in order to learn end-to-end mappings from corrupted images to the ground truths. Whilst the convolution being the feature extractor while eliminating the noise, deconvolution aims to recover the image details. The authors propose to introduce skip connections between each convolution layer and the corresponding deconvolution layer for the back-propagation of  gradients to lower layers and passing  image details to upper layers. The authors claim  "setting new records" on image denoising and super-resolution, which is somehow exaggerated. 
\begin{figure}[htbp]
\centering
\includegraphics[width=15cm]{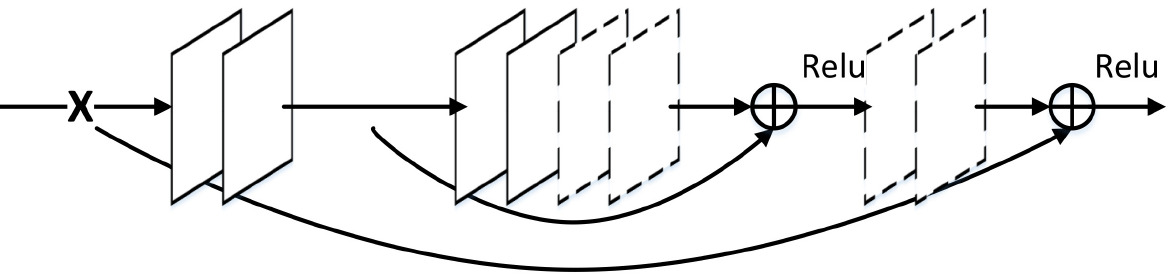}\\
\caption{RED-Net~\cite{Mao2016}.}
\label{fig:RED-Net}
\end{figure}
The method is claimed to be compared with half a dozen methods, including SRCNN. The results are comparable as only a slight improvement is reported on the test cases in hand; there is no mention of time efficiency. It can be deduced from the results that increasing the number of layers improve the results slightly but the efficiency part is still elusive and does not seem cost effective. Moreover, as illustrated in Fig.~\ref{fig:RED-Net}, the method consists of $n$ convolutional layers followed by $n$ deconvolutional layers and there is a skip connection between every $k$th convolution layer and $(n-k+1)$th deconvolution layer, for $k=1,2,\ldots,n$. In other words, the first convolution layer must wait the last deconvolution layer for the sake of correct correspondence. For such a deep network, this may be an additional overhead.

A method presented in~\cite{Mao2016a}, to treat partially pixelated images for super-resolution, is named Depixelated Super Resolution Convolutional Neural Network (DSRCNN). It consists of an autoencoder inspired by~\cite{Makhzani2015} combined with two depixelate layers (de-noising and encoding) via deconvolution as illustrated in~\cite{Xu2014}. The autoencoder is composed of a generator and a discriminator. The generator is modeled on SRCNN-Ex~\cite{Dong2014} and consists of de-noising, encoding and decoding layers. 
Application of the method, on randomly pixelated images, shows results comparable to SRCNN but the reported sample size is too small. Moreover the work suffers from clarity of presentation, especially on the placement of depixelate layers; are these part of autoencoder or outside?

\begin{figure}[htbp]
\centering
\includegraphics[width=15cm]{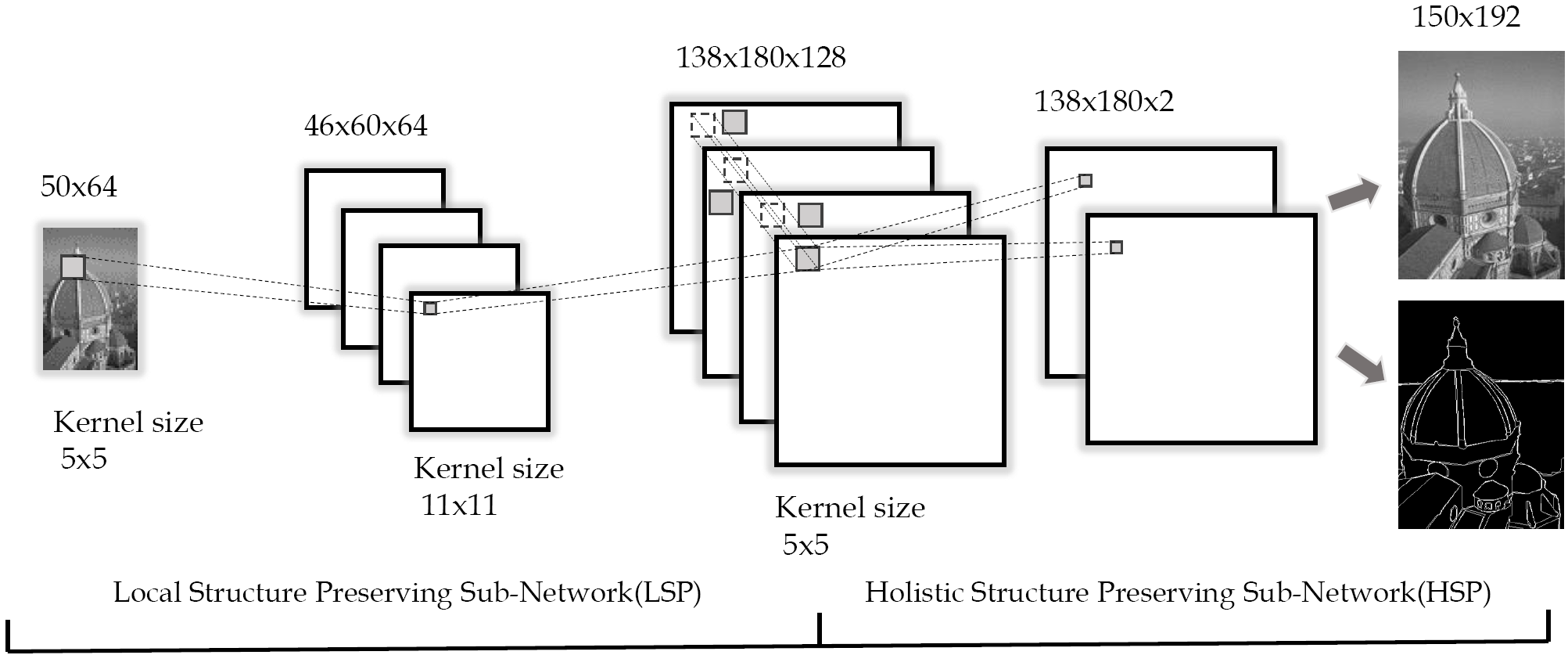}\\
\caption{LSP/HSP network~\cite{Shi2016a}.}
\label{fig:Shi}
\end{figure}
A yet to be properly published work~\cite{Shi2016a} argues against treating all pixels equally and call for taking into account the salient structures in the form of local and holistic contents. With that in view, a local structure-preserving sub-network (LSP), followed by a holistic structure-preserving sub-network (HSP), are proposed to be incorporated into the fully-convolutional learning. By using deconvolution, the LSP upsamples the low resolution patches which are thereafter refined through convolution by HSP, as shown in Fig.~\ref{fig:Shi}. The authors claim superior results in comparison to state of the art methods especially the SRCNN. The paper, in its present form however, suffers from the effective presentation of the results.
The authors criticize SRCNN for its equal treatment of pixels while ignoring the texture contrast and its use of bicubic interpolation in its pre-processing step for it may have adverse effects on the main structure if not initialized properly. 

For single text-image Super-Resolution, the authors in~\cite{Peyrard2015} compare two methods after essential tweaking, namely the Multi-Layer Perceptron (MLP) method reported in~\cite{Pan2003} and one of the pioneering works on CNN or ConvNets~\cite{LeCun1998}. Their example-based strategy attempt to learn a non-linear mapping between pairs of text patches and high-frequency coefficients. According to their results, the CNN based method outperformed the other and a few other contemporary methods they used for comparison. Later in the article, however, the authors extend their experiments to SRCNN, in which case the latter fared superior; still their basic conclusion remains unaltered, i.e. CNN based methods fare far better.
The work in~\cite{Peyrard2016} carries out SR without any prior information on the blurring kernel (hence the word "Blind"in the title) by using CNNs. The authors do some limited simulations on Set5 and Set14 in comparison to SRCNN, A+ and SRF~\cite{Schulter2015}. The claimed results are better than not only in a blind set-up but also with the same reference methods in a non-blind environment; one exception is the superior performance of SRF on Set5.

\begin{figure}[htbp]
\centering
\includegraphics[width=15cm]{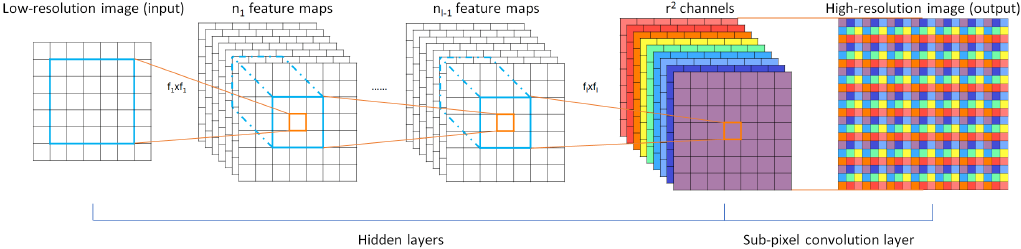}\\
\caption{ESPCN~\cite{Shi2016}.}
\label{fig:ESPCN}
\end{figure}
For the sake of economizing on space and time complexities, the method in~\cite{Shi2016}\footnote{https://github.com/Tetrachrome/subpixel} first downsamples the HR image, after Gaussian filtering via convolution, to LR and then extracts the feature maps from the latter. Then each layer, excluding the last sub-pixel convolution layer, is characterized by its own upscaling filter for the feature map of the concerned layer. The learning of these filters is, however, the job of the last "sub-pixel convolution layer" which upscales the low resolution image to a super-resolved image. The authors call their method the efficient sub-pixel convolutional neural network (ESPCN) which is illustrated in Fig.~\ref{fig:ESPCN}. Note that, for a total of $L$ layers, the preprocessing bicubic filter of SRCNN has been replaced with $L-1$  upscaling filters, each trained for every feature map. The authors report results that are at least better from state of the art methods (read SRCNN) by +0.15dB with images. As of efficiency, the authors report a run-time average performance of  29.5dB over less than 0.2 seconds in a specific environment as against 29.0dB over more than 1 second for SRCNN, while using the set14 dataset.  

With ESPCN~\cite{Shi2016}, two issues come to fore. One, why an available HR image should be reduced to its low resolution counterpart? One may agree that for testing and learning purposes, the HR images are kept as references (ground truths) under the assumption that they are unknown to the system. But what the author suggests is on the contrary. If the aim is just to reduce the size, why not partition the HR image to small tiles, feed it in lieu of low resolution image to the pipeline to get the super-resolved tile and then stitch or tessellate~\cite{Hayat2010} the output tiles? The second issue is the claim of best real-time performance on the basis of a few datasets. While the reported results are commendable, nothing can substitute theoretical complexity analyzes, which is missing from the article.  In a later explanation~\cite{Shi2016b}, the authors do try to address the first issue, while quoting the works in~\cite{Zeiler2014, Zeiler2011}\footnote{http://caffe.berkeleyvision.org/doxygen/classcaffe\_1\_1DeconvolutionLayer.html}. It must be borne in mind that the quoted works are mainly concerned with classification problems, as against image reconstruction. What is hard to understand that deconvolution is an inverse problem and how can it be equivalent with convolution. 

\begin{figure}[htbp]
\centering
\includegraphics[width=15cm]{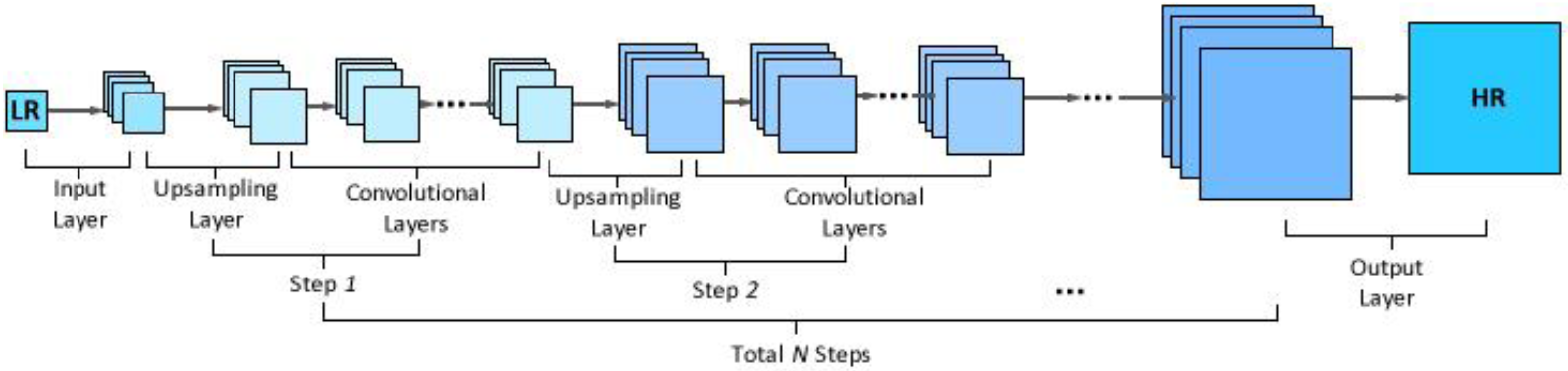}\\
\caption{GUN~\cite{Zhao2017}.}
\label{fig:GUN}
\end{figure}
A yet to be properly published deep CNN based SISR method, the Gradual Upsampling Network (GUN~\cite{Zhao2017}), utilizes a gradual process to magnify from LR to HR, as against the two extremes in SRCNN and ESPCN. The architecture (Fig.~\ref{fig:GUN}) comprises of an input ReLU based convolution layer, followed by a series of alternating upsampling and convolutional layers, and an output layer. The authors believe that their gradual upsampling strategy of adopting very small magnification factor is cost effective in terms of efficiency but no empirical or theoretical evidence is provided. Their training strategy is also gradual and trains an initial network with edge-like samples, followed by gradual tuning with more complex samples. The authors report results that are superior to about nine state of the art methods with VDSR and SRCNN finishing second and third, respectively.

Ren \emph{et al.}~\cite{Ren2015} argue that the operators in CNNs and sparse auto-encoders are translation invariant and are therefore not suited for scenarios requiring translation variant interpolation (TVI). By employing the Shepard interpolation framework~\cite{Shepard1968}, they propose what they call Shepard Convolutional Neural Networks (ShCNN)\footnote{ http://www.deeplearning.cc/shepardcnn and https://github.com/jimmy-ren/vcnn\_double-bladed/tree/master/applications/Shepard\_CNN } in order to introduce and train end-to-end TVI operators in the network, efficiently, while adding a few feature maps in the new Shepard layers. 
The reported super-resolution results, with reference to a few methods including SRCNN and A+, demonstrated superiority of the method. The authors claim 'competitive' running time but no supporting evidence is given, although they mention casually to prefer introducing a few Shepard layers rather than going for a deeper architecture, for efficiency. Secondly, the PSNR results on Sect14 are under-reported for SRCNN.

By employing the Gibbs model as the conditional probability distribution, the authors of~\cite{Bruna2016} initialize their CNN with filters having good geometric properties - e.g. multi-scale complex wavelets based scattering networks~\cite{Bruna2013} or VGG networks~\cite{Cimpoi2015}\footnote{http://www.robots.ox.ac.uk/$\sim$vgg/research/deeptex/}. They also put forward a time-costly fine-tuning algorithm via gradient estimation of the conditional log-likelihood. The CNN architecture is inspired SRCNN. For training simulations, the authors have randomly selected 64x64 image patches from a subset of the training set of ImageNet~\cite{Deng2009}. 
The ensued results are not enviable both in terms of PSNR and Time complexity.

\begin{figure}[htbp]
\centering
\includegraphics[width=15cm]{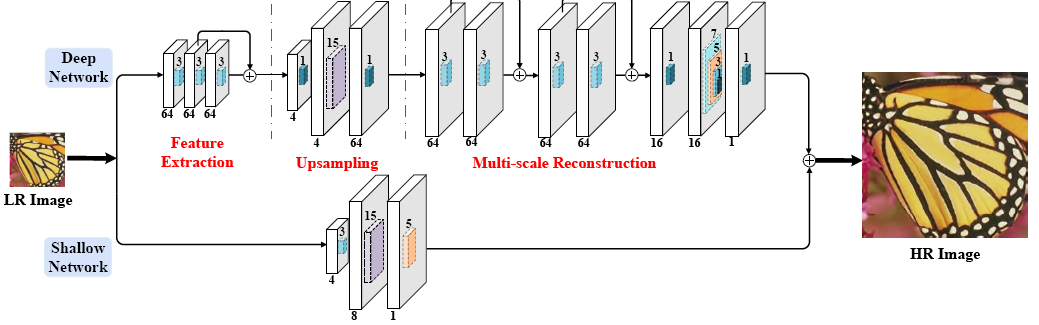}\\
\caption{EEDS~\cite{Wang2016}.}
\label{fig:EEDS}
\end{figure}
The authors of~\cite{Wang2016}argue against the use of bicubic upsampling as a first step and instead propose to go first for feature extraction, in order to map the LR image into a deep feature space, followed by a learning based upsampling of such features to the desired dimensions with learned filters (see Fig.~\ref{fig:EEDS}). For the HR reconstruction, context information is derived from the upsampled features, in a multi-scale way that incorporates both short- and long-range contextual information at the same time. By taking a cue from~\cite{He2016}, the authors introduce a shallow network as part of their architecture, in order to facilitate training, for the sake of efficiency and faster convergence. 
The method is compared with more than ten other methods - including SRCNN, ARFL~\cite{Schulter2015}, NBSRF~\cite{Salvador2015} and CSCN (strangely, not with CSCN-MV from the same work~\cite{Wang2015}) - and is reported to be superior, with CSN being second. The authors also show gradual increase in quality with increase in kernel size for single scale reconstruction, but still the multiscale version outperforms all. Time complexity is given in the form the number of iterations; the EEDS convergence is of the order of $10^5$ iterations, for even a small dataset like set5. It would have been interesting, however, had the EEDS been compared with other methods in this context.

Two inter-related works~\cite{Kato2016,Ohtani2017}, with the former being a specific case of the latter with $k=2$, employ a $k\times k$ -channel CNN with the output pixels, having the same coordinates, grouped together. The result is a magnified image, with an scaling factor of $k$. The method does not involve any bicubic interpolation. 
The authors report average PSNR gain of about 0.2dB par rapport the SRCNN across all the upscaling factors of 2,3 and 4. In ~\cite{Kato2016}, they claim an improvement of 0.39 dB (the average PSNR 36.88 dB) over SRCNN for an upscaling factor of 2 but none of SRCNN works~\cite{Dong2016,Dong2014} have reported any such average PSNR (36.49dB). 

\begin{figure}[htbp]
\centering
\includegraphics[width=15cm]{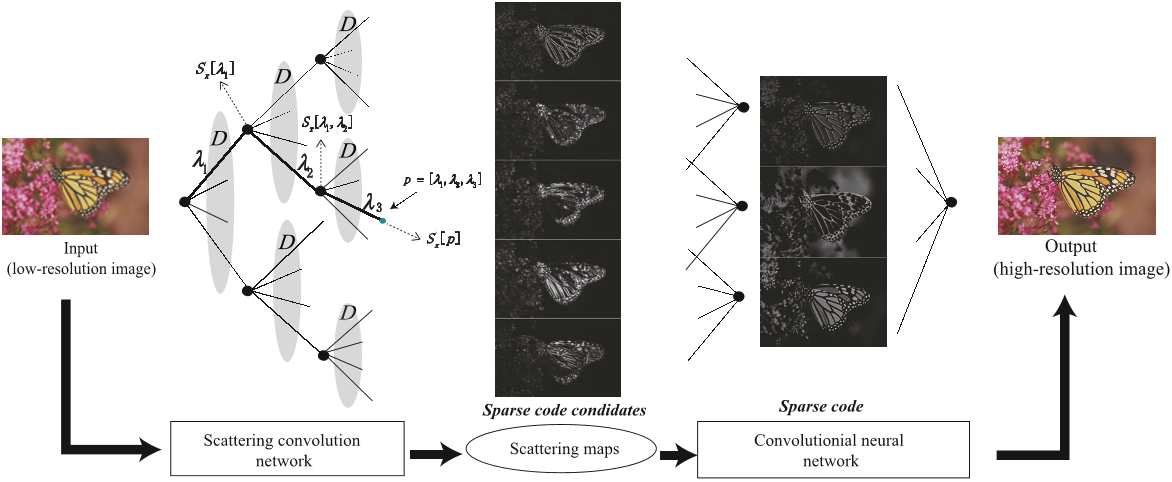}\\
\caption{HWCN~\cite{Gao2016}.}
\label{fig:HWCN}
\end{figure}
In ~\cite{Gao2016}, the LR image is input to a scattering convolution network~\cite{Mallat2012,Bruna2013} to get scattering maps from which sparse codes are extracted which serve as input to a CNN. The method is referred to by the authors as hybrid wavelet convolution network (HWCN) as it is a m\'elange of both sparse coding and CNN, as shown in Fig.~\ref{fig:HWCN}. Given a tiny dataset, HWCN could train complex deep network with better generalization by regularization from scattering convolutions, and thereby is a competitive alternative to CNNs. 
Although the reported results outperform sparse-coding methods by a small margin but no comparison is given par rapport the CNN methods. Our own reading of the comparison with SRCNN, suggests that the latter outperforms it. As of time complexity of convergence, only comparison with the bicubic interpolation is given. The convergence time is reported to be in the order 1000 iterations, but the convergence to a PSNR of around 34dB is a bit dubious?

The multi-channel-input SR convolutional neural network (MC-SRCNN~\cite{Youm2016}) takes LR and its various enhanced interpolated versions (hence the term multichannel) as inputs to extend the SRCNN. With this the expectation is to extract better features for HR restoration as the interpolated versions of LR may have complementary information to that of SRCNN exclusive architecture. With an upscaling factor of 3 and 18-channel MC-SRCNN, the authors report an improvement of 0.34dB and  0.18dB in PSNR on set5 and set14, respectively.  

In~\cite{Osendorfer2014}, both CNN and sparse coding are employed whereby the input LR image is subjected to an efficient convolutional sparse coding module followed by 'perforated' upsampling and convolutional decoding. The 'perforated' upsampling may bring more sparsity and somehow approximates inverse max-pooling operator of deep CNNs. For comparison, the upsampling part is improvised, to give various reference methods, with classical interpolation methods, like bilinear, bicubic, nearest neighbor and linear shift. The authors call such variations as state of the art methods, which is a bit naive. Anyhow a PSNR gain of about 0.8dB, over simple bicubic interpolation, may not be helpful in establishing the effectiveness of the method.
According to~\cite {Shi2016}, such an approach of increasing image resolution, in the middle of the network gradually, may escalate the computational complexity which may be especially problematic with CNNs where the processing speed is a function of the input image resolution. 

In~\cite{Riegler2015}, the authors improvise some contemporary state of the art super-resolution methods~\cite{Schulter2015,Dong2014,Timofte2015,Timofte2013} using conditioned regression models in order to exploit supplementary kernel information during training and inference. The idea is  to have a single training model, rather than repeating it for every candidate blur kernel, especially if the latter is different for each image. In the proposed "Regression-conditioned" SRCNN, the first convolutional layer is replaced with parametrized convolution in the form of a non-linear function derived via an additional neural network trained jointly with SRCNN. In the ensued results, however, the conditioned SR forests outperform conditioned SRCNN by a small margin.

As a part of their work in~\cite{Johnson2016}, the authors train their CNN with semantics in the form of feature reconstruction loss rather than the per pixel loss of SRCNN. In their results, the subjective quality is reported to be far superior to SRCNN, but quantitative metrics do not support the claim. The authors attribute it to "the feature reconstruction loss" that leads "to a slight cross-hatch pattern visible under magnification, which harms its PSNR and SSIM". 
A SISR method~\cite{Qu2016} employ CNN in combination with regularization constraints involving both the local similarity and non-local image similarities. The authors claim a state-of-the-art resolution quality. 
The work in~\cite{Tang2016} take into account both the local intensity and local gradient to reduce edge blurring while training the SRCNN model.  

Whilst CNN had been the preferred choice of the researchers, other kinds of deep networks have got little attention. In the presentation of SR generative adversarial network (SRGAN~\cite{Ledig2016}\footnote{Links to implementations:\\https://github.com/leehomyc/Photo-Realistic-Super-Resoluton\\https://github.com/junhocho/SRGAN\\https://github.com/titu1994/Super-Resolution-using-Generative-Adversarial-Networks}), the authors claim photo-realistic natural images with $\times 4$ magnification after proposing a perceptual loss function constituted by both an adversarial loss (uses a discriminator network to differentiate photorealistic images from SR images) and a content loss that depends on perceptual similarity rather than pixel similarity. The reported PSNRs (27.02dB on Set14 which is even less than bicubic interpolation) are not that encouraging but the perceptual quality is enviable.

The work in~\cite{Brosch2015} deals with the problem of efficient training of convolutional deep belief networks by learning the weights in the frequency domain in order to avoid the time-expensive convolutions. The authors claim about $\times 8$ efficiency on 2D images and $\times 200$ on 3D volumes from medical data.

Another deep learning method, the Laplacian Pyramid Super-Resolution Network (LapSRN~\cite{Lai2017}) attempts to progressively reconstruct the sub-band residuals of HR images. At each pyramid level, the input coarse-resolution feature maps which are used to predict the high-frequency residuals and subsequent transposed convolutions for upsampling to the finer level. The authors claim both efficiency and accuracy par rapport the state of the art methods, especially SRCNN.
\subsection{Set14-Based Benchmarking}
In Table~\ref{table:benchSet14}, based on the application of a given method to Set14 dataset, we have collected the data from the literature on PSNR of final super-resolved image with an upscaling factor of $3$. The reason to select Set14 and upscaling factor of $3$ is the popularity of this benchmark with most the presentations. We tried to standardize the empirical time complexity but every work has used its own experimental settings and objective analysis may not be possible in the absence of theoretical complexity analysis. 
We will not pass any verdict based on a small dataset for a particular scaling factor and leave it to the reader. 
\begin{table} [htb]
\renewcommand\tabcolsep{2pt}
\caption{Benchmarking over Set14 with $\times 3$ upscaling.}
\label{table:benchSet14}
\centering
\begin{tabular}{|c|l|l|}
\hline
 \textbf{S/No.} & \textbf{Name with reference} & \textbf{PSNR (dB)}\\ 
\hline 
 1 & Bicubic &27.54\\
 2 & SC~\cite{Yang2010} & 28.31\\
 3 & K-SVD~\cite{Zeyde2012} & 28.67\\
 4 & ANR~\cite{Timofte2013} & 28.65\\
 5 & A+~\cite{Timofte2015} & 29.13\\
 6 & NE+LLE~\cite{Chang2004} & 28.60\\
 7 & KK~\cite{Kim2010} & 28.94\\
 8 & SRCNN-Ex~\cite{Dong2014} & 29.00\\
 9 & SRCNN~\cite{Dong2016} & 29.30\\
 10 & FSRCNN~\cite{Dong2016a} & 29.43\\
 11 & DPN~\cite{Liang2017a} & 29.80\\
 12 & ShCNN~\cite{Ren2015} & 29.39\\
 13 & Self-Ex~\cite{Huang2015b} & 29.16\\
 14 & VDSR~\cite{Kim2016} & 29.77\\
 15 & DRCN~\cite{Kim2016a} & 29.76\\
 16 & HWCN~\cite{Gao2016} & 29.17\\
 17 & SCN~\cite{Wang2015} & 29.41\\
 18 & CSCN~\cite{Liu2016} & 29.55\\
 19 & MSCN-4~\cite{Liu2017} & 29.65\\
 20 & RED30~\cite{Mao2016} & 29.61\\
 21 & DSRCNN~\cite{Mao2016a} & 28.60\\
 22 & NBSRF~\cite{Salvador2015} & 29.25\\
 23 & R-basic~\cite{Liang2017} & 29.67\\
 24 & R-deep~\cite{Liang2017} & 29.80\\
 25 & RFL~\cite{Schulter2015} & 29.05\\
 26 & ARFL~\cite{Schulter2015} & 29.13\\
 27 & RFL+~\cite{Schulter2015} & 29.17\\
 28 & ARFL+~\cite{Schulter2015} & 29.23\\
 29 & ESPCN~\cite{Shi2016} & 29.49\\
 30 & IA~\cite{Timofte2016} & 29.69\\
 31 & DJSR~\cite{Wang2015a} & 29.96\\
\hline
\end{tabular}
\end{table}

Note that we have not included GUN~\cite{Zhao2017} in the table as the authors have re-executed all their benchmarking methods by applying these to Y channel and doing simple bicubic interpolation to the CbCr components, as against resorting to RGB. Their PSNR results on Set14 at $\times 3$ are therefore on the high side, i.e. in dBs; 33.35 (GUN), 33.07 (VDSR), 32.93 (SRCNN), 32.39(A+) and even 30.36 (bicubic). The authors of DJSR~\cite{Wang2015a} report 0.3dB higher PSNR for SRCNN which may be a typo.  
\subsection{Importance of SRCNN}
All the comparisons notwithstanding, SRCNN is still a landmark work on SR. Not only being a quintessential benchmarking method, it has also been employed in many application scenarios. 
In a work on face recognition~\cite{Rasti2016}, the authors got improved recognition rates when the input face images were subjected to SRCNN-Ex~\cite{Dong2014} before the use of HMM and SVD.
In~\cite{Zhang2016}, SRCNN is successfully employed to improve the quality of infrared thermography (IRT) images for object recognition.
In~\cite{Huang2016}, the authors  compare the performance of SRCNN with another SR method~\cite{Huang2006} in improving the resolution of lensless blood cell counting images; SRCNN is reported to have 9.5\% better results.
The method in~\cite{Simo-Serra2016} can process sketch images of any resolution and employs CNNs to auto-clean rough raster sketch drawings.
The authors of~\cite{Yoon2015} apply CNN to Light-Field (LF) images - hence the name LFCNN - to upsample both the angular and spatial resolutions. 
\section{Video Super-resolution}\label{sec:VSR}
\subsection{Video Benchmarks}
\begin{table} [htb]
\begin{minipage}{\textwidth}   
\renewcommand\tabcolsep{2pt}
\caption{Publicly available video datasets.}
\label{table:video_datasets}
\begin{tabular}{|c|p{5cm}|p{13cm}|}
\hline
 \textbf{S/No.} & \textbf{Name with reference} & \textbf{Details}\\ 
\hline 
 1 & Middlebury\footnote{http://vision.middlebury.edu} & Has many datasets classified to five categories, like for optical flow~\cite{Baker2011}.\\
 2 & Harmonic Inc.\footnote{https://www.harmonicinc.com/free-4k-demo-footage/} & Arrays of 4K (Ultra HD) demo footage, like Myanmar 60p~\cite{Kappeler2016}, in  H.264 or ProRes 422 HQ format with 50p or 60p samples.\\
3 & Videoset4 or Vid4\footnote{https://twitter.box.com/v/vespcn-vid4} & Consists of four test videos - \emph{walk, foliage, city,} and \emph{calendar} - which were also used in~\cite{Liu2014}\\
4 & Xiph.org Test Media\footnote{https://media.xiph.org/} & A collection of test sequences and video clips, e.g. Derf's collection\footnote{https://media.xiph.org/video/derf/}\\
5 & Ultra Video Group\footnote{http://ultravideo.cs.tut.fi/} & A variety of 4K 120fps test sequences in raw as well as other state of the art formats, like HEVC,  in cooperation with Digiturk\footnote{http://www.digiturk.com.tr/}.\\
6 & YFCC100M \cite{Thomee2015} & As already described elsewhere, has 0.8 million are videos.\\
7 & VidSet12~\cite {Greaves2016} & Following the guidelines of~\cite {Liao2015} constructed a test set of 48 sequences from 12 HR videos, each 31 frames long.\\
8 & VideoSR\footnote{http://www.cse.cuhk.edu.hk/leojia/projects/DeepSR/}~\cite{Liao2015} & 160 video sequences from 26 high-quality 1080p HD video clips of a variety of scenes and objects. 
\\
9 & CDVL\footnote{ http://www.cdvl.org/} & Consumer Digital Video Library (CDVL) contains 115 uncompressed full HD videos excluding repeated videos.\\
10 & Sintel.~\cite{Butler2012, Wulff2012} & A computer generated dataset; popular for optical flow development. \\
11 & 25 YUV format video sequences\footnote{http://www.codersvoice.com/a/webbase/video/08/152014/130.html}~\cite{Liu2014} & The available link works no more.\\
\hline
\end{tabular}
\end{minipage}
\end{table}
Table~\ref{table:video_datasets} lists some of the popular video datasets that are publicly available. Note that still image datasets are still valid and are mainly employed to check the quality of single frames.

Not only the video specific methods, but also the popular image SR methods described in Section~\ref{sec:img} - especially SRCNN, A+, ESPCN and bicubic interpolation - are usually employed for comparison of video SR works. In addition, following are the non-CNN video SR methods preferred for benchmarking:
\begin{enumerate}
	\item \textbf{3DSKR~\cite{Takeda2009}:} adaptive enhancement and spatio-temporal scaling, without explicit motion estimation, by solving a local weighted least-squares problem, where the weights are derived from the space/time comparison of neighboring pixels.
	\item \textbf{ANN~\cite{Cheng2012}:} employs artificial neural network (ANN) to learn spatio-temporal detail between LR and HR frames.
	\item \textbf{BayesSR~\cite{Liu2014}:} a Bayesian strategy to adaptively carry out HR frames reconstruction while at the same time also estimating motion, blur kernel and noise.
	\item \textbf{Bayesian-MB~\cite{Ma2015}:} an expectation maximization (EM) strategy that specially focuses motion blur (MB) by optimally searching least blurred pixels for residual blur estimation and HR reconstruction.
\end{enumerate}
A few works have benchmarked against Video Enhancer~\cite{VideoEnh20xx}, a commercially available software. 

\subsection{Video Super-Resolution}
The ESPCN~\cite{Shi2016}, already described elsewhere, had been shown to perform well (+0.39dB) on 1080p videos with 0.03--0.04s per frame execution time as against 0.434 for SRCNN.

\begin{figure}[htbp]
\centering
\includegraphics[width=15cm]{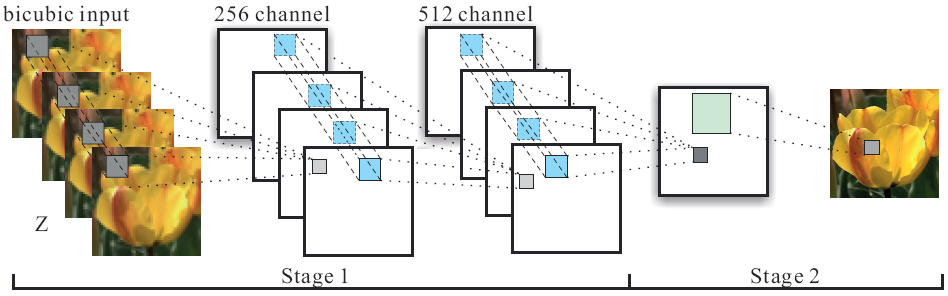}\\
\caption{The CNN part of DraftCNN~\cite{Liao2015}.}
\label{fig:DraftCNN}
\end{figure}
As a variation to the iterative Bayesian adaptive multi-frame SR (MFSR) strategy~\cite{Liu2014} of maximum \emph{a posteriori} (MAP) optical flow, noise level, and blur kernel estimation, the work in~\cite{Liao2015} proposes a non-iterative framework that generates a draft-ensemble from LR frame sequence, followed by the employment of a CNN to determine the optimal draft. The framework\footnote{ http://www.cse.cuhk.edu.hk/leojia/projects/DeepSR/} has thus two parts, \emph{viz.} a) the feed-forward SR draft ensemble generation and b) a deep CNN to non-linearly combine the drafts (Fig.~\ref{fig:DraftCNN}). Note that a SR draft ensemble is "the set of high-resolution patch candidates before final image deconvolution." 
The method is reported to have outperformed the reference method~\cite{Liu2014}, as far as video quality is concerned. Time complexity has not been mentioned.

The video SR method (VSRnet) in~\cite{Kappeler2016} uses motion compensated consecutive frames as input to a CNN\footnote{http://ivpl.eecs.northwestern.edu/software}. Following a three layer SRCNN architecture, the method offers three alternatives 
to combine consecutive video frames, i.e. concatenation a) before layer 1, b) between layer 1 and 2 or c) between layer 2 and 3. For the sake of  accuracy and speed, the authors train their system on images (what they call pre-training) and then employ the ensued filter coefficients to initialize the video training. In addition, the authors claim $20\%$ more time efficiency by employing what they call Filter Symmetry (in temporal sense) Enforcement (FSE); by this they mean equal weightage to t-i and t+i filter, if t represent the central frame temporally. Their motion compensation strategy is flexible enough to deal with motion blur, in the case of fast moving objects. A comparison, with four image based (including A+ and SRCNN) and four video based methods, has been reported and the method is shown to be superior to all at upscale factors of 2 and 3; with upscale factor of 4, it is outperformed by BayesSR~\cite{Liu2014}.
As far as time efficiency is concerned, VSRnet demonstrate marginally better execution time (accompanied with even better PSNR) in comparison to the four video-based methods (ANN, Bayesian-MB, BayesSR and Video Enhancer) but it is still slow for obvious reasons associated with videos. In addition, the video dataset is small (Mayanmar sequence with 59 scenes out of which 6 for training plus 4 from other sources they call Videoset4). They offer image pre-training as an alternate solution to the the creation of a large video database; a strategy also adopted in~\cite{Kappeler2016a}. Still an image cannot be taken as a replacement for a video.

In~\cite{Kappeler2016a}, VSRnet is tweaked for compressed videos; hence the name CVSRnet.  In contrast to a typical video super-resolution method, rather than taking the compression information - like frame type or quantizer step - from the encoder, CVSRnet relies on the compressed LR frames for the reconstruction of HR video. The rest of the strategy is the same as VSRnet; even the same eight methods have been used for comparison with almost similar results. 

A strategy, outlined in~\cite{Gupta2009}, generates super-resolved video frames from LR videos and periodically realizes HR still frames. The original idea was to have a separate sensor in the video camera for periodically acquiring stills while recording a low resolution video; these stills were to be used in refining the videos. Zheng~\cite{Zheng2016} carries out a CNN based implementation of this strategy, with the input being  a sequences of video frames in which the first and last frames are HR, while the rest are LR. The HR frames are warped using the CNN based optical flow scheme, called FlowNet~\cite{Dosovitskiy2015}, wherein the optical flow, from the first and last frames to each of the middle frame (LR), is calculated. The warped HR frames are subjected to a Graph-cut composition along with the upsampled low resolution frames and concatenated thereafter to produce a super-resolved video.
The implementation~\cite{Zheng2016} suffers from efficiency in comparison to SRCNN. This may be attributed to the expensive nature of FlowNets~\cite{Dosovitskiy2015}. The implementer puts the blame on the time spent on data transit. Moreover, with Sintel database they used, the results are not enviable which may be due to the lacking of natural details because of the synthetic nature of the dataset and the author feels that there is less information to discard during downsampling. 

\begin{figure}[htbp]
\centering
\includegraphics[width=15cm]{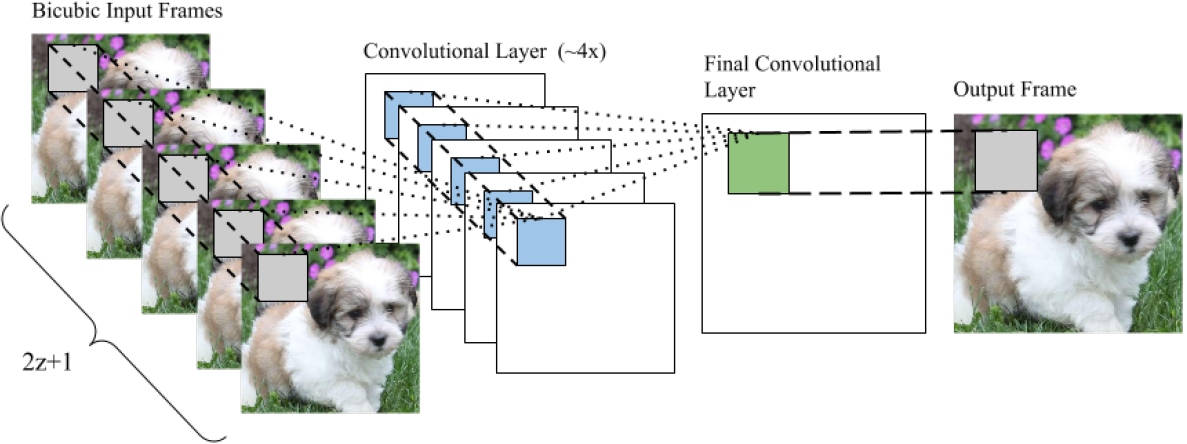}\\
\caption{MFCNN~\cite{Greaves2016}.}
\label{fig:MFCNN}
\end{figure}
The  multi-frame (MFCNN) video SR method in~\cite{Greaves2016} improves the resolution of a given frame based on the pixel values in the adjacent frames, within distance $d$ on either side. In other words, information from a total of $2d+1$ frames is concatenated along the channel dimension of the CNN, as shown in Fig.~\ref{fig:MFCNN}. The authors employ a SRCNN inspired architecture, with $9$ ReLU based layers and dropout, to train on still images (SICNN), followed by single frames (SFCNN) and ultimately extending to multi-frames (MFCNN). Testing with Set14 databases shows that SICNN has comparable results to SRCNN for obvious reason, but both are outperformed by CSCN-MV~\cite{Wang2015}. But the video version (MFCNN) is reported to have outperformed DraftCNN and BayesSR on their own set of videos (VidSet12). The authors claim minimal data pre-processing and computation cost but themselves recognize inconsistencies in results in certain situations, e.g. LR scenes with trees and leaves.
The authors of~\cite{Greaves2016} are hopeful about the potential role of RCNN (R for recurrent) in video super-resolution.

\begin{figure}[htbp]
\centering
\includegraphics[width=15cm]{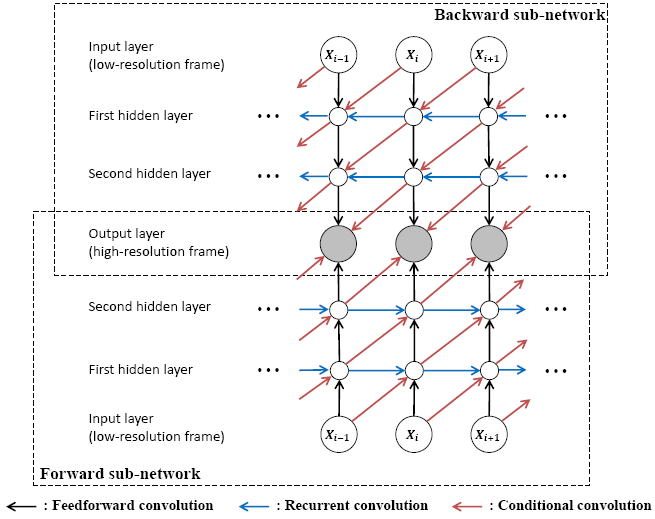}\\
\caption{BRCN~\cite{Huang2015}.}
\label{fig:BRCN}
\end{figure}
In~\cite{Huang2015}, a bidirectional recurrent CNN (BRCN) is proposed for MFSR, under the belief that a recurrent neural network (RNN) are better suited to "model long-term contextual information of temporal sequences". As illustrated in Fig.~\ref{fig:BRCN}, the method employs three types of convolutions, \emph{viz.} a) feed-forward convolution for LR to HR spatial correspondence, b) recurrent convolution for learning temporal dependence via weightage based linking of hidden layers of adjacent frames, and c) conditional convolution to connect previous inputs to current hidden layer in order "to enhance visual-temporal dependency". Empirically, the method is reported to have been slower than methods, like ANR and SRCNN, but with better visual quality (read PSNR); theoretical complexity analysis is, however, lacking.

\begin{figure}[htbp]
\centering
\includegraphics[width=15cm]{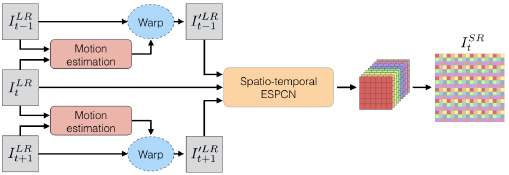}\\
\caption{VESPCN~\cite{Caballero2017}.}
\label{fig:VESPCN}
\end{figure}
The video version of ESPCN (VESPCN~\cite{Caballero2017}), illustrated in Fig.~\ref{fig:VESPCN}, relies on spatio-temporal sub-pixel CNNs that take into account the temporal redundancies and a fast multi-resolution spatial transformer motion compensation module. Although the authors analyze a) early fusion (as employed in~\cite{Kappeler2016}), b) slow fusion~\cite{Karpathy2014} and 3) 3D convolutions~\cite{Tran2015} for the joint processing of multiple consecutive video frames; the experiments are restricted to early and slow fusions only. The authors boast a computational cost economy of 30\% par rapport SFMSR techniques; a 0.2dB PSNR gain if computational cost is kept constant. The benchmarking results on Vid4 dataset establishes the superiority of VESPCN against VSRnet, ESPCN and SRCNN, both in terms of visual quality and time efficiency.

An action recognition method~\cite{Nasrollahi2015} is reported to have improved results by enhancing the video quality via SRCNN before subjecting it to the recognition method proposed in~\cite{Wang2013}.

\subsection{Videoset4-Based Benchmarking}
Table~\ref{table:benchVid4} compares various SR methods on the basis of Videoset14 at an upscaling factor of $\times 3$. Note that the first five are image SR methods. For the table we have relied on two sources only~\cite{Kappeler2016,Caballero2017}, as most of the works are inconsistent in their approach as far as the datasets are concerned. One can count VSRnet as lossless CVSRnet and obviously lossy CVSRnet will give lower PSNR. ESPN, which is mainly image based method, is reported to have outperformed VSRnet in the presentation of VESPN ( an extension of ESPN by almost the same team); even the reported PSNR is lesser (26.64 dB) than shown in the table (26.79 dB). For the effectiveness of the rest of the methods one needs to consult individual works. Strangely enough, in the presentation of MFCNN~\cite{Greaves2016}, bicubic interpolation outperforms not only BayesSR but also the single frame version of the method (SFCNN) on the VidSet12 dataset. The results of the work concerning BRCN relied on a dataset which is no more accessible, at least at the time of writing of this article. 
\begin{table} [htb]
\renewcommand\tabcolsep{2pt}
\caption{VSR Benchmarking over Vid4 with $\times 3$ upscaling.}
\label{table:benchVid4}
\centering
\begin{tabular}{|c|l|l|}
\hline
 \textbf{S/No.} & \textbf{Name with reference} & \textbf{PSNR (dB)}\\ 
\hline 
 1 & Bicubic & 25.28\\
 2 & SC~\cite{Yang2010} & 26.01\\
 3 & A+~\cite{Timofte2015} & 26.36\\
 4 & SRCNN~\cite{Dong2016} & 26.51\\
 5 & ESPCN~\cite{Shi2016} & 26.97\\
 6 & ANN~\cite{Cheng2012} & 25.94\\
 7 & BayesSR~\cite{Liu2014} & 25.82\\
 8 & Bayesian-MB~\cite{Ma2015} & 26.43\\
 9 & Video Enhancer~\cite{VideoEnh20xx} & 26.34\\
 10 & VSRnet~\cite{Kappeler2016} & 26.79\\
 11 & VESPCN~\cite{Caballero2017} & 27.25\\
\hline
\end{tabular}
\end{table}

\section{Depth Maps/3D and higher dimensions}\label{sec:3D}
\subsection{Benchmarks}
Table~\ref{table:depths} lists some of the important publicly available datasets. Unlike, the datasets from the previous sections, these datasets mostly address specific situations, as described in the third column of the table.   
\begin{table} [htb]
\begin{minipage}{\textwidth}   
\renewcommand\tabcolsep{2pt}
\caption{Publicly available 3D sequences and depth maps.}
\label{table:depths}
\centering
\begin{tabular}{|c|p{5cm}|p{13cm}|}
\hline
 \textbf{S/No.} & \textbf{Name with reference} & \textbf{Details}\\ 
\hline 
 1 & Middlebury stereo dataset 2001~\cite{Scharstein2002}, 2003~\cite{Scharstein2003}, 2005/06~\cite{Hirschmuller2007}, and 2014~\cite{Scharstein2014} & contains HR textures and depths with lots of details~\cite{Song2017}.\\
 2 & Laser Scan.~\cite{Aodha2012}  & Specially developed for patch based SR.\\
3 & Sintel~\cite{Butler2012, Wulff2012} & a synthesized data-set via physical simulations that contains lots of depth details and high quality images~\cite{Song2017}.\\
4 & SENTINEL-2\footnote{For details on Sentinel-1 to 5 and 5P visit https://sentinels.copernicus.eu/web/sentinel/sentinel-data-access/access-to-sentinel-data} & images having 13 channels (as against 3 in RGB) with a ground resolution of up to 10m, and a high radiometric resolution (more than 8 bit per pixel for each channel).\\
5 & ICL-NUIM~\cite{Handa2014} & For benchmarking RGB-D, Visual Odometry and SLAM algorithms with two different scenes, \emph{viz.} the living room and the office room scene.\\
6 & NYU Depth~\cite{Silberman2012} & Composed of 464 indoor video sequences from a Microsoft Kinect camera; 249 scenes for training and 215 for testing.\\ 
7 & KITTI~\cite{Geiger2013} & composed of several outdoor scenes captured while driving with car-mounted cameras and depth sensor.\\
8 & ToFMark~\cite {Ferstl2013} & Time-of-Flight (ToF) Depth Upsampling Evaluation Dataset having LR ToF depth acquisition together with a HR intensity image.\\ 
9 & Open Access Series of Imaging Studies (OASIS~\cite {Marcus2007}) & A collection of cross-sectional MRI Data from 416 subjects aged 18 to 96 year.\\
\hline
\end{tabular}
\end{minipage}   
\end{table}

Aside from some of the methods described in Section~\ref{sec:img} - like ANR, A+, NE+, K-SVD, SRCNN - many of the following non-CNN methods are usually employed as benchmarks: 
\begin{enumerate}
	\item \textbf{Guided Image Sampling~\cite{He2013}:} an edge preserving filter that can be used joint upsampling.
	\item \textbf{MRF~\cite{Diebel2005}:}  Markov Random Fields based enhancement of LR depths with insight from the accompanied HR camera images under the assumption that depth discontinuities are usually in harmony with intensity changes in the associated camera image.  
	\item \textbf{ATGV~\cite{Ferstl2013}:} depth image upsampling guided by an anisotropic diffusion tensor that is  computed from HR intensity image. 
	\item \textbf{3D-ToF Upsampling~\cite{Park2011}:} attempts to super-resolve LR depths from noisy 3D time-of-flight (3D-ToF) camera coupled with a HR RGB camera by using non-local mean filtering to regularize depth maps and a multi-features based edge weighting scheme based on the HR RGB input.
	\item \textbf{PatchSDSR~\cite{Aodha2012}:} increases the depth resolution by matching the height field of each LR patch input against only a generic database of local HR patches; the selection of right HR candidate is done via MRF labeling.
	\item \textbf{Edge-guided~\cite{Xie2016a}:}	super-resolves a single depth image with the help of a HR edge map, obtained by MRF optimization from the edges in its LR counterpart. 
\end{enumerate}
Other important non-CNN  methods for depth SR benchmarking include~\cite{Hornacek2013,Xie2014,Ferstl2015}.
\subsection{Literature on CNN based SR of 3D/Depths and Multispectral Data}
\begin{figure}[htbp]
\centering
\includegraphics[width=15cm]{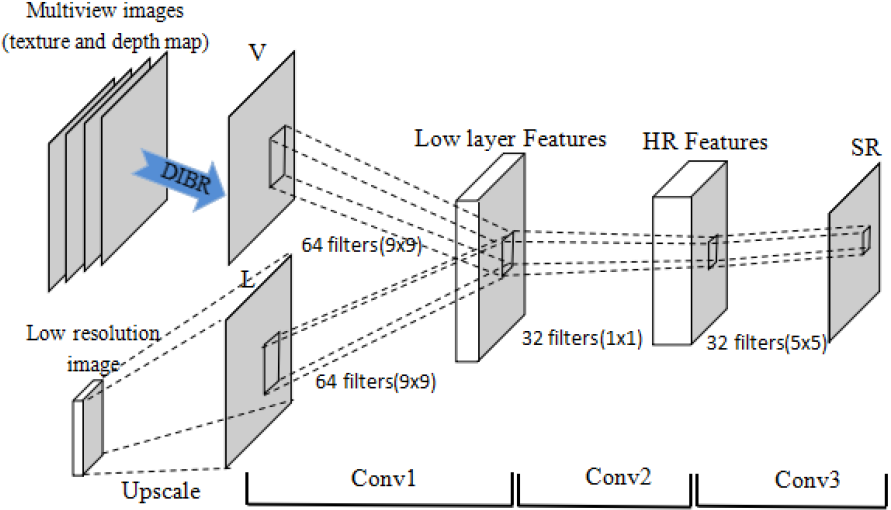}\\
\caption{3DSRCN~\cite{Xie2016}.}
\label{fig:3DSRCN}
\end{figure}
In 3DSRCN~\cite{Xie2016}, the LR viewpoints in the 3D video are up-sampled. The resultant interpolated LR and a projected virtual view are fed to the first layer of a three-layered CNN, as shown in Fig.~\ref{fig:3DSRCN}. In the experimental set up, Layer-1 involves $2\times 64$ filters (9x9), layer-2 involves 64x32 filters (1x1)  and layer-3 contains and 32x1 filters (5X5), all with the convolution stride of 1. Layer-1 combines the two inputs to get a set of feature maps, layer-2 establishes the mapping from low to high resolution features, while layer-3 uses the ensued features to get the super-resolved image. The simulation results, on samples from Middlebury stereo 2014~\cite{Scharstein2014}, report a PSNR gain of about 1dB par rapport the SRCNN and another virtual-view assisted method reported in~\cite{Jin2016}. The authors also claim empirical time efficiency.

\begin{figure}[htbp]
\centering
\includegraphics[width=15cm]{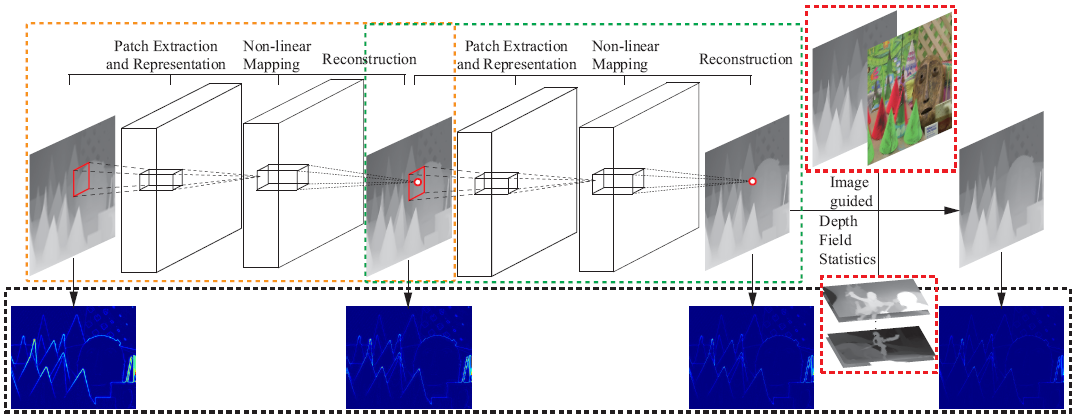}\\
\caption{Progressive Deep CNN~\cite{Song2017}.}
\label{fig:PDCNN}
\end{figure}
The work in~\cite{Song2017} proposes what the authors call a progressive deep CNN structure (Fig.~\ref{fig:PDCNN}) that develop HR depth images by a gradual learning of the higher frequencies. Supplementary information in the form of depth field statistics and color-depth correlation serves as two priors provide complement to the CNN for further refinement of depth maps.The authors argue that the method is equally good in the absence of HR color images whereby the depth images by themselves are enough to refine the depth images. The reported results compare the method with more than a dozen state of the art methods applied to standard datasets and the results establish the superiority of the method.

The single image based depth estimation method in~\cite{Eigen2014} employs two deep networks \emph{viz.} coarse and fine. Whilst the former predicts the global depth of the scene, the latter refines it locally. The output of coarse network also works as supplementary first-layer input of image features to the fine network. The authors exploit a scale-invariant error in measuring the depth relations instead of scale. 
The reported results after application to NYU Depth and KITTI datasets are better than the state of the art methods but do not establish the need to use the fine network as its contribution to the final results needs a cost/benefit analysis.

The DEM super resolution, a term first coined in~\cite{Xu2015} wherein the non-local algorithm focused on some learning examples which were itself derived from the original DEM by dividing the latter to overlapping patches and then categorizing as test dataset and learning dataset based on the fact whether they pertain to HR measurements or not.  By calculating the weighted sums of similar patches, HR DEM were restored. The same team employs a three layers' CNN model~\cite{Chen2016}  to extend this strategy, with special focus on DEM compatibility and robustness, wherein layer 1 detects features from the input, layer 2 integrates the ensued features and layer 3 transforms the integrated features to get the super-resolved DEM. The author claim good results but they are compared with bicubic interpolation only.

In order to get a super-resolved output from a single LR depth map, the method ATGV-Net~\cite{Riegler2016}\footnote{https://griegler.github.io/} attempts to combine CNN with an energy minimization model, in the form of a powerful variational model. They improvise SRCNN to estimate, in addition to refining the depth map, the locations of discontinuities in the HR output depth maps. The refined depths and the located discontinuities serve as input to a variational model wherein pairwise regularization is carried out via anisotropic Total Generalized Variation (TGV)~\cite{Bredies2010}, with weights dependent on the network output. The output is the final HR estimate of the depth map. The authors train their model entirely on their own synthetic depth data. For benchmarking however, they use four different datasets; two derived from Middlebury and one each from Laserscan and ToFMark. The method is demonstrated to show better results in comparison to different state of the art methods, especially~\cite{Ferstl2013} which is shown to be second best.

\begin{figure}[htbp]
\centering
\includegraphics[width=15cm]{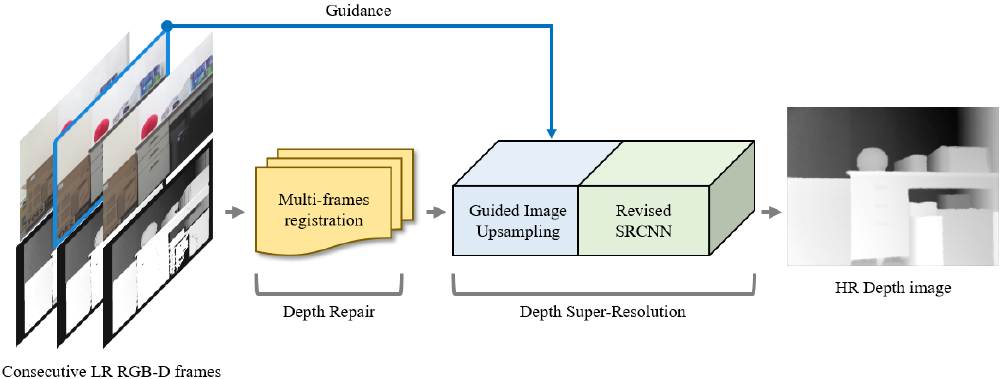}\\
\caption{Depth SR by~\cite{Tseng2016}.}
\label{fig:Tseng}
\end{figure}
The work in~\cite{Tseng2016} first subjects multiple RGB-D images to registration via 3D alignment from 3D feature point correspondences and then employs guided image filtering~\cite{He2013} and a tailored SRCNN in order to obtain HR depth images. The method is illustrated in Fig.~\ref{fig:Tseng}.
The reported results demonstrate the utility of the method when applied to ICL-NUIM~\cite{Handa2014}, MPI-Sintel~\cite{Butler2012} datasets, but adds little to the results of only guided image filtering~\cite{He2013} in the face of additional cost involved in the form of SRCNN.

In~\cite {Moon2016}, the authors employ an already SR trained CNN (transfer learning\footnote{ http://cs231n.github.io/transfer-learning/}) to post-process lunar images in order to estimate the DEM. For the latter part, they introduce to their architecture an additional ConvNet, which is trained anew. The authors have compared the method with bicubic interpolation only which is outperformed for obvious reasons. 

The authors of~\cite{Liebel2016} argue that CNN SR approaches are as useful for remote sensing data as for still images. Their extension of SRCNN to multispectral information (hence the names msiSRCNN) is reported to have given superior results to conventional interpolation methods in terms of PSNR. They had applied the method to SENTINEL-2 images with 13-channel (spectral bands), including the 3 RGB channels, encoded with 16 bit per pixel JPEG2000 format. It would have been, however, better to report RMSE for some distance critical channels. 
A work on 3D cardiac Magnetic Resonance (MR) imaging~\cite{Oktay2016} super-resolves the corresponding 2D planes using an improvised SRCNN that is modified by employing residual learning in LR-HR mapping and using a deeper network with the upscaling being realized learning a deconvolution layer rather than a fixed kernel. 
SRCNN has been utilized to enhance spatial features in the hyperspectral image reconstruction model proposed in~\cite{Li2017}.
Another interesting read on multi-spectral resolution is a three part article on CosmiQ~\cite{Hagerty2017} in the context of SpaceNet challenge\footnote{http://crowdsourcing.topcoder.com/spacenet}. 
As described earlier the authors of~\cite{Brosch2015} claim time efficiency $\times 200$ on 3D volumes on medical data from OASIS.
\subsection{Banchmarking over a standard dataset}
Before dwelling on the comparison, it must be noted that whilst for texture quality PSNR and SSIM are suited, the root mean square error (RMSE) is usually the preferred metric for depth maps. RMSE in taken as a length unit, like meter (m) or its subdivisions and even \emph{pixel disparity}. An analysis of RMSE par rapport the view distance in given in~\cite{hayat2010a}. 
As of comparing various methods, unfortunately, due to their dealings with diverse problems, most of the works have used different approaches of benchmarking, especially in choosing the datasets. Even with the same dataset, the subset chosen is different from one work to another. Then there are inconsistencies in reporting the results. For example, the works in~\cite{Riegler2016,Song2017} have benchmarked their methods on Middlebury and Laser Scan data, with many common reference methods. Just a look at $\times 4$ upscaling part on RMSE, as listed in Table~\ref{table:bench3D}, reveals glaring inconsistencies in their reported results. 
\begin{table} [htb]
\renewcommand\tabcolsep{2pt}
\caption{RMSE results for \emph{Cones} from Middlebury at $\times 4$.}
\label{table:bench3D}
\centering
\begin{tabular}{|c|l|l|l|}
\hline
 \textbf{S/No.} & \textbf{Name with reference} & \textbf{Riegler \emph{et al.}~\cite{Riegler2016}} & \textbf{Song \emph{et al.}~\cite{Song2017}}\\ 
\hline 
 1 & Nearest neighbor & 6:1236 & 6.0054\\
 2 & Bicubic & 4.9544 & 3.8635\\
 3 & ANR~\cite{Timofte2013} & 3.0256 & 3.3156\\
 4 & K-SVD~\cite{Zeyde2012} & 3.8468 & 3.2232\\
 5 & ATGV~\cite{Ferstl2013} & 3.6372 & 3.9968\\
 6 & 3PatchSDSR~\cite{Aodha2012} & 6.0168 & 12.6938\\
\hline
\end{tabular}
\end{table}
\section{Conclusion}\label{sec:concl}
Despite the claims of accuracy and time efficiency, in almost all the works reviewed in this survey, a lot needs to be done on both fronts. The accuracy improvements are marginal and that too with inconsistencies; even there are instances that work A claims superiority to work B and after a while B comes up with superior results after minor tweaking - and this goes on and on. A minor thing to point out is that none of the works have reported the quality of input LR par rapport the HR during training. When it comes to time efficiency, empirical results are commonplace but little can be found on the theoretical front in the form of asymptotic analysis etc. Although there are attempts to exploit sparse coding for deep learning, the race with the former is in full flow without realizing the fact that the issue is SR, not the superiority of one paradigm over other. 

So far most of the attention has gone to CNNs and the potential of other deep learning approaches is yet to be exhaustively investigated. But the most important thing is the loss of sight from  high scaling factors. In~\cite{Ledig2016}, the authors rightly pose the question, "how do we recover the finer texture details when we super-resolve at large upscaling factors?" To them the problem is the thrust on the choice of the objective function and on minimizing the MSE, which may yield high PSNRs but at the expense of losing high-frequency details, thus leading to poor subjective quality of the output HR.

Deep learning has its constraints. According to Shalev-Shwartz \emph{et al.}~\cite{Shalev-Shwartz2017}, it is haunted by failures from  at least four origins, viz. a) non-informative gradients, b) inefficiency of a network left to learn itself, c) architecture choice and conditioning and d) flat activations. A researcher must not be oblivious to these while designing a deep learning network. This discussion is incomplete without an allusion to a news article~\cite{Yao2017} on the limitations of deep learning. It's worth a read.


%


\ifCLASSOPTIONcaptionsoff
  \newpage
\fi



\bibliographystyle{IEEEtran}

\end{document}